\relax
\documentclass[letterpaper]{article} 
\usepackage{aaai22}  
\usepackage{times}  
\usepackage{helvet}  
\usepackage{courier}  
\usepackage[hyphens]{url}  
\usepackage{graphicx} 
\usepackage{natbib}  
\usepackage{caption} 
\DeclareCaptionStyle{ruled}{labelfont=normalfont,labelsep=colon,strut=off} 
\usepackage{algorithm}
\usepackage{algorithmic}
\usepackage{color,xcolor}
\usepackage{newfloat}
\usepackage{listings}
\floatstyle{ruled}
\newfloat{listing}{tb}{lst}{}
\floatname{listing}{Listing}
\usepackage{amssymb}
\usepackage{amsmath}
\usepackage{amsfonts}
\usepackage{amsthm}
\usepackage{dsfont}
\usepackage{multirow}
\usepackage{enumerate}
\usepackage{pifont}
\usepackage{subfigure}

\DeclareMathOperator*{\argmax}{arg\,max}

\newtheorem{Proposition}{Proposition}

\usepackage{booktabs}

\title{Learning Robust Policy against Disturbance in Transition Dynamics\\via State-Conservative Policy Optimization}
\author{
Yufei Kuang\textsuperscript{\rm 1}, 
Miao Lu\textsuperscript{\rm 1}, 
Jie Wang\textsuperscript{\rm 1,2}\thanks{Corresponding author.}, 
Qi Zhou\textsuperscript{\rm 1}, 
Bin Li\textsuperscript{\rm 1}, 
Houqiang Li\textsuperscript{\rm 1,2}
}

\affiliations{
\textsuperscript{\rm 1}CAS Key Laboratory of Technology in GIPAS, 
University of Science and Technology of China\\
\textsuperscript{\rm 2}Institute of Artificial Intelligence, Hefei Comprehensive National Science Center\\
\{yfkuang, lumiao, zhouqida\}@mail.ustc.edu.cn\\
\{jiewangx, binli, lihq\}@ustc.edu.cn
}

\begin{document}

\maketitle

\begin{abstract}
Deep reinforcement learning algorithms can perform poorly in real-world tasks due to the discrepancy between source and target environments. 
This discrepancy is commonly viewed as the disturbance in transition dynamics. 
Many existing algorithms learn robust policies by modeling the disturbance and applying it to source environments during training, which usually requires prior knowledge about the disturbance and control of simulators. 
However, these algorithms can fail in scenarios where the disturbance from target environments is unknown or is intractable to model in simulators. 
To tackle this problem, we propose a novel model-free actor-critic algorithm---namely, \textbf{S}tate-\textbf{C}onservative \textbf{P}olicy \textbf{O}ptimization (SCPO)---to learn robust policies \textit{without modeling the disturbance in advance}. 
Specifically, SCPO reduces the disturbance in transition dynamics to that in state space and then approximates it by a simple gradient-based regularizer. 
The appealing features of SCPO include that it is simple to implement and does not require additional knowledge about the disturbance or specially designed simulators. 
Experiments in several robot control tasks demonstrate that SCPO learns robust policies against the disturbance in transition dynamics.

\end{abstract}

\section{Introduction}

Deep reinforcement learning (DRL) has achieved remarkable success in many complex control tasks \cite{dqn,ddpg, dr, openai-cube}. 
However, existing DRL algorithms tend to perform poorly in real-world tasks due to the environment discrepancy \cite{epopt, mrpo}. 
For example, policies trained to control robots in source environments can fail to generalize in target environments with slightly changed physical parameters (e.g., mass and friction). 
This discrepancy usually comes from test-generalization and
simulation-transfer in real-world tasks, and it is commonly viewed
as the disturbance in transition dynamics \cite{rarl, tessler2019action}.  Therefore, research on learning robust policies against the disturbance in transition dynamics is receiving increasing attention.

In order to learn robust policies, many previous algorithms model the disturbance and apply it to source environments during training.
One simple and effective algorithm is domain randomization (DR) \cite{dr}, which randomizes the parameters in source environments to generate an ensemble of similar transition dynamics. DR uses these dynamics to approximate the potential disturbance in target environments. 
Another line of research is based on robust Markov decision process (RMDP) theory \cite{robust-dp}, which considers the worst-case disturbance in transition dynamics at each step during training. 
For example, 
robust adversarial reinforcement learning (RARL) algorithm \cite{rarl} models the disturbance as trainable forces in source environments based on the insight that environment discrepancy can be viewed as extra forces, and then RARL learns the worst-case disturbance by adversarial training.

Learning robust policies by modeling the disturbance in advance, though effective, has two limitations in practice. First, these algorithms usually require task-specific prior knowledge to model the disturbance \cite{model-misspecification}.
For example, to randomize the environmental parameters during training, DR requires a pre-given uncertainty set that contains specific parameters and their corresponding variation ranges;
to approximate the disturbance in target environments, RARL employs extra forces that are manually designed for each task. 
However, when we train policies for new tasks or test in unseen environments, we may lack enough knowledge about the disturbance \cite{tessler2019action}. 
Second, these algorithms usually assume control of specially designed simulators \cite{benchmarks}, as disturbing the agents during training in the real world can be dangerous. 
However, sometimes building new simulators can be expensive, and sometimes the disturbance is even intractable to model in simulators (e.g., disturbance caused by wear-and-tear errors and aerodynamic effects in robot control tasks) \cite{abdullah2019wasserstein}.
In general, learning robust policies by modeling the disturbance and applying it to source environments can be impractical in many scenarios.

In this paper, we propose a novel algorithm to learn robust policies without modeling the disturbance in advance. 
First, we reduce the disturbance in transition dynamics to that in state space based on the intuition that any disturbance in transition dynamics eventually influences the process via the change of future states. Then, we propose the \textbf{S}tate-\textbf{C}onservative \textbf{M}arkov \textbf{D}ecision \textbf{P}rocess (SC-MDP), which considers the worst-case disturbance in state space at each step.
Compared with the RMDP, SC-MDP reduces the original constrained optimization problem in transition dynamics to a simpler one in state space, whose solving process does not require taking infimum over a subset of the infinite dimensional transition probability space.
Finally, we propose a model-free actor-critic algorithm, called \textbf{S}tate-\textbf{C}onservative \textbf{P}olicy \textbf{O}ptimization (SCPO), which approximates the disturbance in state space by a simple gradient-based regularizer and learns robust policies by maximizing the objective of SC-MDP. 
The appealing features of SCPO include that it is simple to implement and does not require task-specific prior knowledge and specially designed simulators to model the disturbance. 
Experiments in MuJoCo benchmarks show that SCPO consistently learns robust policies against different types of disturbance in transition dynamics.

\section{Related Work} \label{sec: related work}

In this section, we discuss related work that aims to learn robust policies from different perspectives.
Note that we mainly focus on learning robust policies against the disturbance in  transition dynamics in this paper.

\textbf{Robustness against the Disturbance in Transition Dynamics} 
The disturbance in transition dynamics  mainly comes from test-generalization and simulation-transfer \cite{rarl, epopt}. Many previous model-free algorithms learn robust policies by modeling the disturbance and applying it to source environments during training.
\citet{dr} propose domain randomization (DR) to generate environments with different transition dynamics.
\citet{robust-dp} propose robust Markov decision process (RMDP), which maximizes cumulative rewards under the worst-case disturbance.
Based on RMDP, \citet{model-misspecification} propose robust MPO (R-MPO) to learn robust policies with pre-given uncertainty sets of environmental parameters; \citet{mrpo} propose monotonic robust policy optimization (MRPO) with a theoretical performance lower bound;
\citet{abdullah2019wasserstein} propose the Wasserstein robust reinforcement learning (W$\text{R}^2$L) to train the environment parameters with protagonists jointly. 
Compared with these algorithms, SCPO does not require specific priors and control of simulators to model the disturbance in advance.

\textbf{Robustness against the Disturbance in State Observations} 
The disturbance in state observations can originate from sensor errors and equipment inaccuracy \cite{robust-observation}.
\citet{fgsm-rl} learn robust policies by adversarial training with FGSM based attacks, 
and \citet{robust-observation} derive a robust policy regularizer that relates to total variation distance or KL-divergence on perturbed policies.
Note that the disturbance in state observations does not affect the ground-truth states and the underlying transition dynamics. Thus, there are gaps betweem the algorithms designed for these two different tasks \cite{robust-observation}.

\textbf{Robustness against the Disturbance in Action Space} The disturbance in action space can come from controller inaccuracy  and environment noise. 
\citet{openai-cube} propose the random network adversary (RNA) algorithm, which injects noise in action space during training using a neural network with randomly sampled weights; \citet{tessler2019action}  propose probabilistic action robust MDP (PR-MDP) and noisy action robust MDP (NR-MDP) to model two types of disturbance in action space and then solve them by adversarial training. 
Sometimes we also use the disturbance in action space to approximate the unmodeled disturbance in transition dynamics, even though the disturbance in transition dynamics is not necessarily originated from the disturbance in action space. 
We compare SCPO with the action robust algorithm in our experiments.

\section{Preliminaries}

In this section, we briefly introduce the preliminaries.

\textbf{Markov Decision Process (MDP)} An MDP is defined by a tuple $(\mathcal{S}, \mathcal{A}, p, r, \gamma)$. Here $\mathcal{S}  \subset \mathbb{R}^m $ is the state space, $\mathcal{A} \subset \mathbb{R}^n$ is the action space, and $p:\mathcal{S}\times\mathcal{A}\rightarrow \Delta_{\mathcal{S}}$ is the transition dynamic, where $\Delta_{\mathcal{S}}$ is the set of Borel probability measures on ${\mathcal{S}}$. 
The reward function $r:\mathcal{S}\times\mathcal{A}\rightarrow[R_{\min}, R_{\max}]$ is bounded, and $\gamma \in (0,1)$ is the discount factor. Let $\pi: \mathcal{S} \rightarrow \Delta_\mathcal{A}$ be a stationary policy, where $\Delta_\mathcal{A}$ is the set of Borel probability measures on $\mathcal{A}$. 
For convenience, we overload the notation to let $\pi(\cdot|s)$ and $p(\cdot|s,a)$ also denote the probability density function (PDF) without ambiguity.

\textbf{Robust Markov decision process (RMDP)} The RMDP considers the disturbance in transition dynamics at each step. It is defined as a tuple $(\mathcal{S}, \mathcal{A}, \mathcal{P}, r, \gamma)$, where $\mathcal{S}$, $\mathcal{A}$, $r$, $\gamma$ are defined as above and $\mathcal{P}$ is an uncertainty set of possible transitions. Specifically, for each $(s,a)\in\mathcal{S}\times\mathcal{A}$, $\mathcal{P}(s,a)$ is a set of transitions  $p(s^{\prime}|s,a):\mathcal{S}\times\mathcal{A}\mapsto\Delta_{\mathcal{S}}$. When the agent is at state $s$ and takes action $a$, the next state $s^{\prime}$ is determined by one of the transitions in $\mathcal{P}(s,a)$. By definition, the uncertainty set $\mathcal{P}$ described all possible disturbance in transition dynamics. 
RMDP aims to learn a policy that maximizes the expected return under the worst-case disturbance \cite{robust-mdp}. The objective of RMDP is defined as
\begin{align}\label{equ:rmdp_objective}
    J_\mathcal{P}(\pi)\triangleq\inf_{p\in\mathcal{P}}\mathbb{E}_{p,\pi}[\sum_{t=0}^{+\infty}\gamma^tr(s_t,a_t)].
\end{align}
That is, taking minimum over all possible probability measures induced by $\pi$ and $\mathcal{P}$ on the trajectory space \cite{robust-dp}.
Based on the objective, the robust value functions are defined as $V_\mathcal{P}^{\pi}(s)\triangleq\inf_{p\in\mathcal{P}}\mathbb{E}_{p,\pi}[\sum_{t=0}^{+\infty}\gamma^tr(s_t,a_t)|s_0=s]$ and 
$Q_\mathcal{P}^{\pi}(s,a)\triangleq\inf_{p\in\mathcal{P}}\mathbb{E}_{p,\pi}[\sum_{t=0}^{+\infty}\gamma^tr(s_t,a_t)|s_0=s,a_0=a]$.
Note that $Q^{\pi}_{\mathcal{P}}$ is the fixed point of the following robust Bellman operator (which is a contraction on $\mathbb{R}^{|\mathcal{S}\times\mathcal{A}|})$
\begin{align}\label{equ:robust_bell}
    \begin{aligned}
    \mathcal{T}_{\mathcal{P}}^{\pi}Q\triangleq r(s,a)
    +\gamma\inf_{p(\cdot|s,a)\in\mathcal{P}(s,a)}\mathbb{E}_{s^{\prime}\sim p(\cdot|s,a),a^{\prime}\sim \pi(\cdot|s^{\prime})}[Q(s^{\prime},a^{\prime})].
    \end{aligned}
\end{align}
Correspondingly, the optimal policy and     the optimal action-value function is defined to be $Q^{\pi^{\star}}_{\mathcal{P}}
\triangleq \sup_{\pi}Q^{\pi}_{\mathcal{P}}$.

\textbf{Wasserstein Uncertainty Set} A bounded uncertainty set makes the problem well-defined, as optimizing over arbitrary dynamics only results in non-performing system.
The uncertainty set $\mathcal{P}(s,a)$ has recently been described using the Wasserstein distance by many researchers \cite{abdullah2019wasserstein,hou2020robust}.
Let $p_0(s^{\prime}|s,a)$ denote the transition probability in the source environment, which is assumed to be fixed. Then, the Wasserstein uncertainty set is defined as
\begin{align}\label{equ:wass_uncer_set}
    \begin{aligned}
        \mathcal{P}_\epsilon \triangleq \Big\{p:\mathbb{W}^{(p)}\big(p(\cdot|s,a), p_0(\cdot|s,a)\big)<\epsilon,\,\,\forall (s,a)\Big\},
    \end{aligned}
\end{align}
where $\mathbb{W}^{(p)}$ is the $p^{\mathrm{th}}$-order Wasserstein metric on $\Delta_{\mathcal{S}}$ given by
$\mathbb{W}^{p}(\mu, \nu)\triangleq(\inf _{\gamma \in \Gamma(\mu, \nu)} \int_{\mathcal{S} \times \mathcal{S}} d(x, y)^{p} \mathrm{~d} \gamma)^{1/p}$, $d$ is a distance metric on $\mathcal{S}$, and $\Gamma(\mu,\nu)$ refers to all the couplings of $\mu$ and $\nu$. 
By recent progresses in optimal transport \cite{blanchet2019quantifying}, the robust Bellman operator (\ref{equ:robust_bell}), when described by Wasserstein uncertainty set (\ref{equ:wass_uncer_set}), can be rewrite without explicitly taking infimum over a set of probability measures. We show the result in the following proposition, which serves as a motivation to our method.

\begin{Proposition}\label{pro:wass_robust_bell}
When the uncertainty set is described by Wasserstein distance of order $p$ as in Equation (\ref{equ:wass_uncer_set}), the robust Bellman operator (\ref{equ:robust_bell}) is equivalent to
\begin{align}\label{equ:wass_robust_bell}
    \begin{aligned}
        \mathcal{T}_{\mathcal{P}_{\epsilon}}^{\pi}Q(s,a)=r(s,a)+\gamma\sup_{\lambda\geq 0}\mathbb{E}_{\tilde{s}\sim p_0(\cdot|s,a)}
        \Big\{\inf_{s^{\prime}\in\mathcal{S}}\mathbb{E}_{a^{\prime}\sim\pi(\cdot|s^{\prime})}[Q(s^{\prime},a^{\prime})]+\lambda\big (d(s^{\prime},\tilde{s})^p-\epsilon\big)\Big\}.
    \end{aligned}
\end{align}
\end{Proposition}

\section{From RMDP to State-Conservative MDP}\label{sec:state_disturbance}

In this section, we propose a new state-conservative objective and the corresponding state-conservative Markov decision process (SC-MDP). SC-MDP reduces the disturbance in transition dynamics to that in state space, allowing us to learn robust policies without full prior knowledge about the disturbance. Before we formally propose the new objective, we first explain the motivation from state disturbance.

\subsection{A View from State Disturbance}\label{subsec:idea}
Consider the robust Bellman operator  (\ref{equ:robust_bell}) and its Wasserstein variant (\ref{equ:wass_robust_bell}). 
We can write the empirical version of (\ref{equ:wass_robust_bell}) as
\begin{align}
    \label{equ:emp_wass_bell}
    \widehat{\mathcal{T}}^{\pi}_{\mathcal{P}_{\epsilon}}Q(s_t,a_t)=r(s_t,a_t)+\gamma\sup_{\lambda\geq 0}
    \left\{\inf_{s^{\prime}\in\mathcal{S}}\mathbb{E}_{a^{\prime}\sim\pi(\cdot|s^{\prime})}\left[Q(s^{\prime},a^{\prime})\right]+\lambda\left( d\left(s^{\prime},s_{t+1}\right)^p-\epsilon\right)
    \right\}
\end{align}
where $s_{t+1}\sim p_0(\cdot|s_t,a_t)$. Now assume that both the value function $V(s^\prime) \triangleq \mathbb{E}_{a^{\prime}\sim\pi(\cdot|s^{\prime})}[Q(s^{\prime},a^{\prime})]$ and the $p^\text{th}$-order distance metric $d(s^{\prime},s_{t+1})^p$ are convex functions of the variable $s^{\prime}$ in Equation (\ref{equ:wass_robust_bell}). By the duality theory \citep{boyd2004convex}, Equation (\ref{equ:emp_wass_bell}) is equivalent to the following form (detailed proof in Appendix A.1):
\begin{align}\label{equ:emp_wass_bell_dual}
    \widehat{\mathcal{T}}^{\pi}_{\mathcal{P}_{\epsilon}}Q(s_t,a_t)=r(s_t,a_t)+
    \gamma \inf_{s^{\prime}\in B_{\epsilon}(s_{t+1})}\mathbb{E}_{a^{\prime}\sim\pi(\cdot|s^{\prime})}[Q(s^{\prime},a^{\prime})].
\end{align}
Such a form involves taking infimum of the future state-value $V^\pi(s^{\prime})$ over all $s^{\prime}$ in the $\epsilon$-ball $B_\epsilon(s_{t+1})\triangleq\{s^{\prime}\in\mathcal{S}:d(s^{\prime},s_{t+1})^p\leq\epsilon\}$. Note that the convexity of $V(s^{\prime})$ can hold under linear and deterministic assumptions on rewards, transitions and policies when the MDP is of finite horizon. 

In general, the state-value function  $V^\pi(s^{\prime})$ is not necessarily convex. However, this observation gives us a different view on promoting robustness against the disturbance in transition dynamics. Intuitively, any disturbance in transition $p$ finally influences the process via the change of future states. Thus, instead of considering the worst-case perturbed transition as in the robust Bellman operator (\ref{equ:robust_bell}), we can directly consider the worst-case perturbed next states as in Equation (\ref{equ:emp_wass_bell_dual}). In other words, we can learn robust policies by considering the worst state-value over \emph{all possible states} close to $s_{t+1}\sim p_0(\cdot|s_t,a_t)$. We refer to this method as \emph{state disturbance}, which is more compatible in practice since it only involves finding a constraint minimum in the finite-dimensional state space and thus does not require priors to model the disturbance in advance. 
Starting from this view, we propose a new objective in Section \ref{subsec:state_conservative_objective} and further extend it to a continuous control RL algorithm in Section \ref{sec:sc_alg}.

\subsection{State-Conservative Markov Decision Process}\label{subsec:state_conservative_objective}
Now we formally propose the \emph{state-conservative objective} based on the state disturbance method in Section \ref{subsec:idea}. 
Different from the objective of RMDP in (\ref{equ:rmdp_objective}), 
we define it as
\begin{align}\label{equ:state_conservative_objective}
    \begin{aligned}
    J_{\epsilon\text{-}\mathcal{S}}(\pi)\triangleq&\mathbb{E}_{s_0\sim d_0}[\inf_{s_0^{\prime}\in B_{\epsilon}(s)}\mathbb{E}_{a_0^{\prime}\sim \pi(\cdot|s_0^{\prime})}[r(s_0^{\prime},a_0^{\prime})
    +\gamma\mathbb{E}_{s_1\sim p_0(\cdot|s_0^{\prime},a_0^{\prime})}[\inf_{s_1^{\prime}\in B_{\epsilon}(s_1)}
    \mathbb{E}_{a_1^{\prime}\sim\pi(\cdot|s_1^{\prime})}[r(s_1^{\prime},a_1^{\prime})\\
    &+\gamma\mathbb{E}_{s_2\sim p_0(\cdot|s_1^{\prime},a_1^{\prime})}
    [\inf_{s_2^{\prime}\in B_{\epsilon}(s_2)}
    \mathbb{E}_{a_2^{\prime}\sim\pi(\cdot|s_2^{\prime})}[r(s_2^{\prime},a_2^{\prime})
    +\cdots]\cdots]
    \end{aligned}
\end{align}
where $d_0$ and $p_0$ represent the initial distribution and the transition dynamic in source environments, $\epsilon$ is a non-negative real controlling the degree of state disturbance, and $B_{\epsilon}(s)\triangleq\{s^{\prime}\in\mathcal{S}:d(s^{\prime},s)\leq\epsilon\}$ is the $\epsilon$-ball in $\mathcal{S}\subseteq\mathbb{R}^m$ induced by a distance metric $d:\mathcal{S}\times\mathcal{S}\mapsto\mathbb{R}$. 
Note that by setting $\epsilon$ to $0$, our objective recovers the original reinforcement learning objective. We call the Markov decision process aiming to maximize the objective (\ref{equ:state_conservative_objective}) as 
state-conservative Markov decision process (SC-MDP). 

\begin{figure}[!h]
    \centering
    \subfigure[An example MDP.]{
        \label{fig: visual-mdp}
        \includegraphics[height=0.15\textwidth]{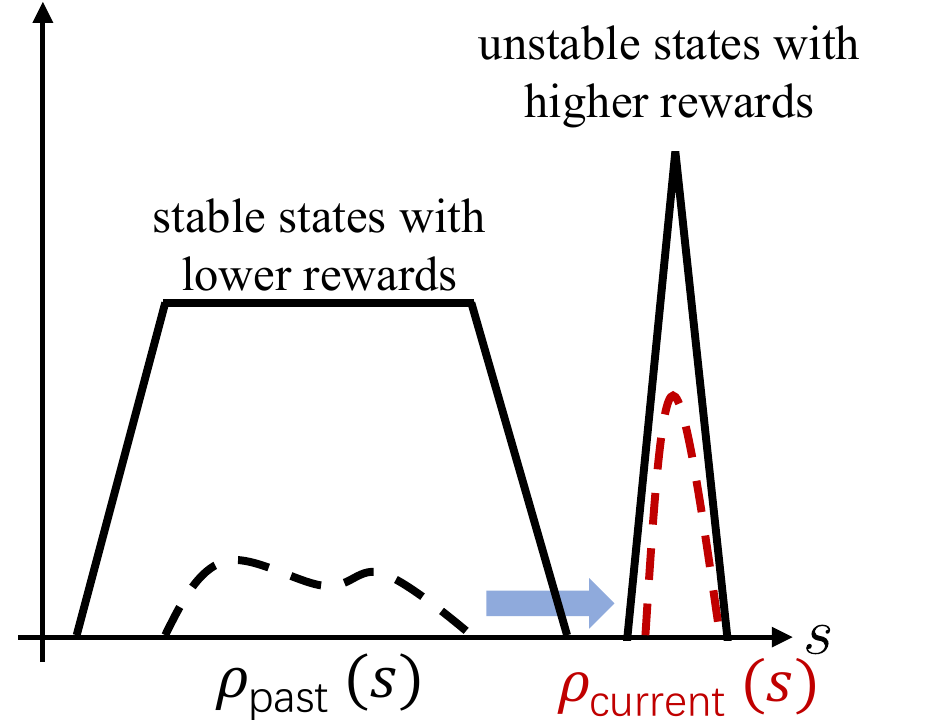}}
    \subfigure[Illustrate the SC-MDP.]{
        \label{fig: visual-sc-mdp}
        \includegraphics[height=0.15\textwidth]{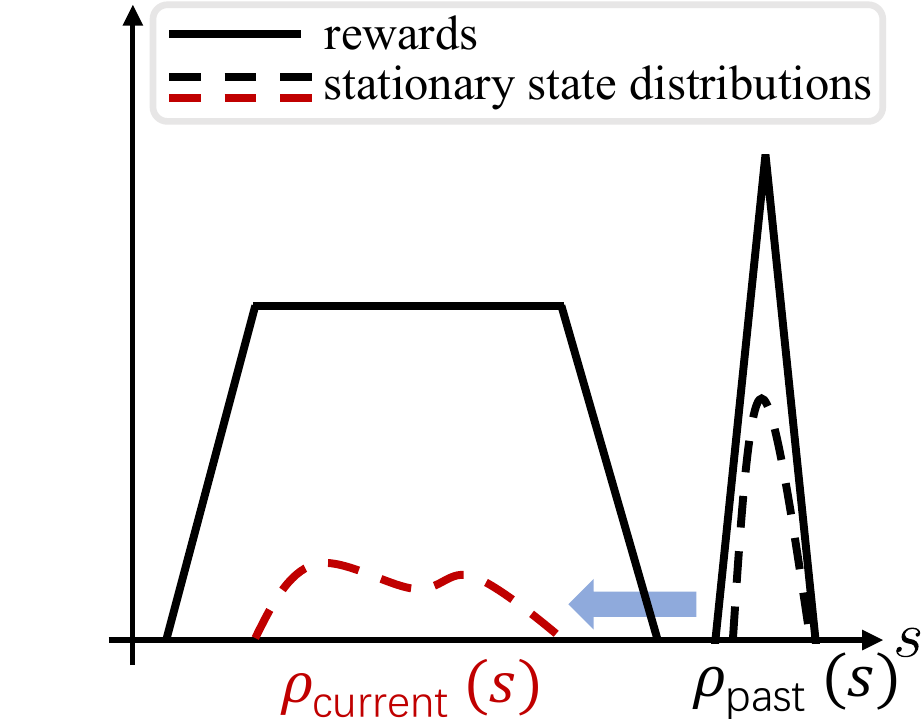}
    }
\caption{An illustration of the state-conservative MDP.}
\label{illustration}
\end{figure}

\textbf{Understand SC-MDP} We illustrate the intuition of SC-MDP in Figure \ref{illustration}. Consider an MDP with one-dimensional state space. Suppose its rewards only depend on $s$. Then, the original objective (i.e., the cumulative rewards) in Figure \ref{fig: visual-mdp} encourages the stationary state distribution $\rho(s)$ to concentrate on states with higher rewards, while the state-conservative objective in Figure \ref{fig: visual-sc-mdp} encourages $\rho(s)$ to concentrate on states that are more stable under disturbance.

\textbf{Relation to RMDP} Our newly proposed SC-MDP coincides with RMDP in specific settings. If we only consider deterministic transitions and let the uncertainty set of RMDP bounded by Wasserstein distance, then one can easily check that our state-conservative objective (\ref{equ:state_conservative_objective}) equals to the objective (\ref{equ:rmdp_objective}) of RMDP.
Though this equivalence does not necessarily hold in general (discussions in Appendix B.1), the new objective helps us design algorithms to learn robust policies.

\subsection{State-Conservative Policy Iteration}\label{sec: SC-PI}
Given a fixed policy $\pi$, we aim to evaluate its value iteratively. Based on the newly proposed objective (\ref{equ:state_conservative_objective}), we define the corresponding action-value function by 
\begin{align}\label{equ:sc_q}
    \begin{aligned}
    Q^{\pi}_{\epsilon\text{-}\mathcal{S}}(s,a)\triangleq& r(s,a)+\gamma\mathbb{E}_{s_1\sim p_0(\cdot|s,a)}[
    \inf_{s_1^{\prime}\in B_{\epsilon}(s_1)}\mathbb{E}_{a_1^{\prime}\sim\pi(\cdot|s_1^{\prime})}[
    r(s_1^{\prime},a_1^{\prime})\\
    &+\gamma\mathbb{E}_{s_2\sim p_0(\cdot|s_1^{\prime},a_1^{\prime})}
    [
    \inf_{s_2^{\prime}\in B_{\epsilon}(s_2)}
    \mathbb{E}_{a_2^{\prime}\sim\pi(\cdot|s_2^{\prime})}[
    r(s_2^{\prime},a_2^{\prime})
    +\cdots]\cdots].
    \end{aligned}
\end{align}
We call $Q^{\pi}_{\epsilon\text{-}\mathcal{S}}$ the state-conservative action-value function of policy $\pi$. It satisfies the following Bellman equation
\begin{align}\label{equ:Bellman}
Q^{\pi}_{\epsilon\text{-}\mathcal{S}}&(s,a)=r(s,a)+
    \gamma \mathbb{E}_{\tilde{s}\sim p_0(\cdot|s,a)}
    \left[\inf_{s^{\prime}\in B_{\epsilon}(\tilde{s})}\mathbb{E}_{a^{\prime}\sim\pi(\cdot|s^{\prime})}[Q_{\epsilon\text{-}\mathcal{S}}^{\pi}(s^{\prime},a^{\prime})]\right].
\end{align}
Thus, $Q^{\pi}_{\epsilon\text{-}\mathcal{S}}$ is the fixed point of the following operator
\begin{align}\label{equ:state_robust_bell}
    \mathbb{T}_{\epsilon}^{\pi}Q&(s,a)=r(s,a)+
    \gamma \mathbb{E}_{\tilde{s}\sim p_0(\cdot|s,a)}
    \left[\inf_{s^{\prime}\in B_{\epsilon}(\tilde{s})}\mathbb{E}_{a^{\prime}\sim\pi(\cdot|s^{\prime})}[Q(s^{\prime},a^{\prime})]\right].
\end{align}
We refer to $ \mathbb{T}_{\epsilon}^{\pi}$ as the state-conservative Bellman operator.
Now we show that $ \mathbb{T}_{\epsilon}^{\pi}$ is a contraction mapping.

\begin{Proposition}[Contraction Mapping]\label{pro:contraction} 
For any $\epsilon\geq 0$ and any fixed policy $\pi$, the state-conservative Bellman operator $ \mathbb{T}_{\epsilon}^{\pi}$ in (\ref{equ:state_robust_bell}) is a contraction mapping on $(\mathbb{R}^{|\mathcal{S}\times\mathcal{A}|},\|\cdot\|_{\infty})$. 
Thus, $ \mathbb{T}_{\epsilon}^{\pi}$ has a unique fixed point, which is just the $Q^{\pi}_{\epsilon\text{-}\mathcal{S}}$.
\end{Proposition}

The proof of Proposition \ref{pro:contraction} is in Appendix A.2. Based on it, we can derive a policy evaluation procedure, which starts from arbitrary $Q^0\in\mathbb{R}^{|\mathcal{S}\times\mathcal{A}|}$ and obtain $Q^{\pi}_{\epsilon\text{-}\mathcal{S}}$ by iteratively calculating $Q^{k+1}=\mathbb{T}^{\pi}_{\epsilon}Q^k$. 
We call this procedure \emph{state-conservative policy evaluation} (SC-PE). 
After evaluating the old policy, we introduce the policy improvement step, which greedily chooses policies state-wisely to maximize the state-conservative objective starting from given states.

\begin{Proposition}[Policy Improvement]\label{pro:improvement}
    Suppose $Q^{\pi_{\text{old}}}_{\epsilon\text{-}\mathcal{S}}$ is the action-value function of $\pi_{\text{old}}$ and $\pi_{\text{new}}$ is chosen greedily by:
    \begin{align}
        \pi_{\text{new}}(\cdot|s)\triangleq\argmax_{\pi\in\Delta_{\mathcal{A}}}\inf_{s^{\prime}\in B_{\epsilon}(s)}\mathbb{E}_{a^{\prime}\sim \pi}[Q^{\pi_{\text{old}}}_{\epsilon\text{-}\mathcal{S}}(s^{\prime},a^{\prime})]
    \end{align}
Then $Q^{\pi_{\text{new}}}(s,a)\geq Q^{\pi_{\text{old}}}(s,a)$ for all $(s,a)\in\mathcal{S}\times\mathcal{A}$.
\end{Proposition}

The proof of Proposition \ref{pro:improvement} is in Appendix A.3.
Now we can derive a policy iteration algorithm that alternates between the state-conservative policy evaluation and improvement steps introduced above. We call it \emph{state-conservative policy iteration} (SC-PI) and show it in Algorithm 1 in Appendix B.2.
By setting $\epsilon$ to $0$, SC-PI recovers the original policy iteration algorithm  \citep{sutton2018reinforcement}.

\section{State-Conservative Policy Optimization}\label{sec:sc_alg}

In this section, we extend the SC-PI algorithm to a model-free actor-critic algorithm, called state-conservative policy optimization (SCPO), in continuous action space. 
First, we introduce the main components of SCPO. Then, we propose a simple and efficient gradient-based implementation of SCPO. Finally, we apply SCPO to the soft actor-critic (SAC) algorithm \cite{sac} and present the state-conservative soft actor-critic (SC-SAC) as an instance.

\subsection{State-Conservative Policy Optimization}\label{sec:scpo}
Let $Q_\theta$ denote the Q-function parameterized by $\theta$, 
$\pi_\phi$ denote the policy parameterized by $\phi$, and $\mathcal{D}$ denote the collected data. 
Then, the state-conservative policy optimization (SCPO) algorithm mainly includes a state-conservative policy evaluation step to \textit{train the critic} and a state-conservative policy improvement step to \textit{train the actor}. We conclude the pseudo code of SCPO in Algorithm 2 in Appendix B.3.

\textbf{State-Conservative Policy Evaluation} In this step, we train the critic $Q_\theta$  by minimizing the Bellman residual
    \begin{align}\label{equ:critic}
    \!\!\!\!J_{\epsilon\text{-}Q}(\theta)\triangleq\mathbb{E}_{(s_t,a_t)\sim \mathcal{D}}\Big[\frac{1}{2}\big(Q_\theta(s_t,a_t)-\widehat{Q}(s_t,a_t)\big)^2\Big],
    \end{align}
where the learning target $\widehat{Q}$ is defined as
\begin{align}
    \widehat{Q}(s_t,a_t) &\triangleq r(s_t,a_t) +  \gamma\mathbb{E}_{s_{t+1}\sim p_0}\big[
     \widehat{V}(s_{t+1}) \big],\\
    \label{equ: target-v}\widehat{V}(s_{t+1}) &\triangleq \inf_{s^{\prime}\in B_{\epsilon}(s_{t+1})}\mathbb{E}_{a^{\prime}\sim\pi_\phi (\cdot|s^{\prime})}\big[Q_{{\theta}}(s^{\prime},a^\prime)\big].
\end{align}

\textbf{State-Conservative Policy Improvement} In this step, we train the actor $\pi_\phi$ by maximizing the objective of SC-MDP
    \begin{align}\label{equ:actor}
    &J_{\epsilon\text{-}\pi}(\phi)\triangleq\mathbb{E}_{s_t\sim D}\Big[\inf_{s\in B_\epsilon(s_t)}\mathbb{E}_{a\sim\pi_\phi(\cdot|s)}\big[Q_\theta(s,a)\big]\Big].
    \end{align}

\textbf{The $\infty$-Norm Based $\epsilon$-Ball.}
We choose the $\infty$-norm as the distance metric on $\mathcal{S}$ to specify $B_{\epsilon}(s)$ in SCPO, i.e.,
\begin{align}\label{equ:ball}
B_{\epsilon}(s)\triangleq\{s^{\prime}\in\mathcal{S}:\|s^{\prime}-s\|_\infty\leq\epsilon\}.
\end{align}
Note that the $\infty$-norm is commonly used to measure noise and disturbance as it indicates that \textit{the intensity of the disturbance in each dimension of the state space $\mathcal{S}$ is independent}. Besides, $\infty$-norm based $\epsilon$-ball allows us to propose a simple and efficient implementation of SCPO in the next section.

\subsection{Gradient Based Implementation}\label{sec: GBI}
To implement the SCPO algorithm proposed in Section \ref{sec:scpo}, we need to solve the following constrained optimization problem which appears in Equation (\ref{equ:critic}) and Equation (\ref{equ:actor}),
\begin{align}\label{equ:imple_target}
    \inf_{s^{\prime}\in B_{\epsilon}(s)}\mathbb{E}_{a^{\prime}\sim\pi_\phi(\cdot|s^{\prime})}[Q_{\theta}(s^{\prime},a^{\prime})].
\end{align}
A natural idea to solve Equation (\ref{equ:imple_target}) is to use the stochastic gradient descent (SGD) method in $B_{\epsilon}(s)$. However, both  performing SGD repeatedly for each batch and estimating $\mathbb{E}_{a\prime\sim\pi(\cdot|s^{\prime})}$ by sampling actions from $\pi(\cdot|s^{\prime})$ are computationally expensive in practice.  To tackle this problem, we propose a gradient based method that \emph{approximately} solves the constrained optimization problem (\ref{equ:imple_target}) \emph{efficiently}.

\textbf{Gradient Based  Regularizer} 

For simplicity, we define ${U}_{\theta,\phi}(s) \triangleq \mathbb{E}_{a\sim\pi_\phi(\cdot|s)}[Q_{\theta}(s,a)]$. By Taylor's Theorem, for given $s\in\mathcal{S}$, we can expand ${U}_{\theta,\phi}(s^{\prime})$ in $B_{\epsilon}(s)$ as 
\begin{align} {U}_{\theta,\phi}(s)+\langle \nabla_{s} {U}_{\theta,\phi}(s),s^{\prime}-s\rangle+o(\|s^{\prime}-s\|_2).
\end{align}
Note that when the disturbance $\epsilon$ is small enough, we can use $\bar{U}_{\theta,\phi,s}(s^{\prime})={U}_{\theta,\phi}(s)+\langle \nabla_{s} {U}_{\theta,\phi}(s),s^{\prime}-s\rangle$ to approximate $U_{\theta,\phi}$ without losing too much information. Then, rather than solving problem (17) directly, we can solve $\inf_{s^{\prime}\in B_{\epsilon}(s)}\bar{U}_{\theta,\phi,s}(s^{\prime})$ instead, which enjoys the following closed form solution under a small $\infty$-norm ball $B_{\epsilon}(s)$,
\begin{align}\label{equ:closed_solution}
\begin{aligned}
    &\inf_{s^{\prime}\in B_{\epsilon}(s)}\bar{U}_{\theta,\phi,s}(s^\prime)={U}_{\theta,\phi}(s)-\epsilon\|\nabla_{s}{U}_{\theta,\phi}(s)\|_{1}.
\end{aligned}
\end{align}
We refer to Equation (\ref{equ:closed_solution}) as a \textbf{G}radient \textbf{B}ased \textbf{R}egularizer (GBR), which approximates the solution to Equation (17) via simply subtracting a gradient norm term.

\begin{figure}[!t]
    \centering
    \subfigure[Deterministic policy.]{
        \label{fig: deterministic}
        \includegraphics[height=0.2\textwidth]{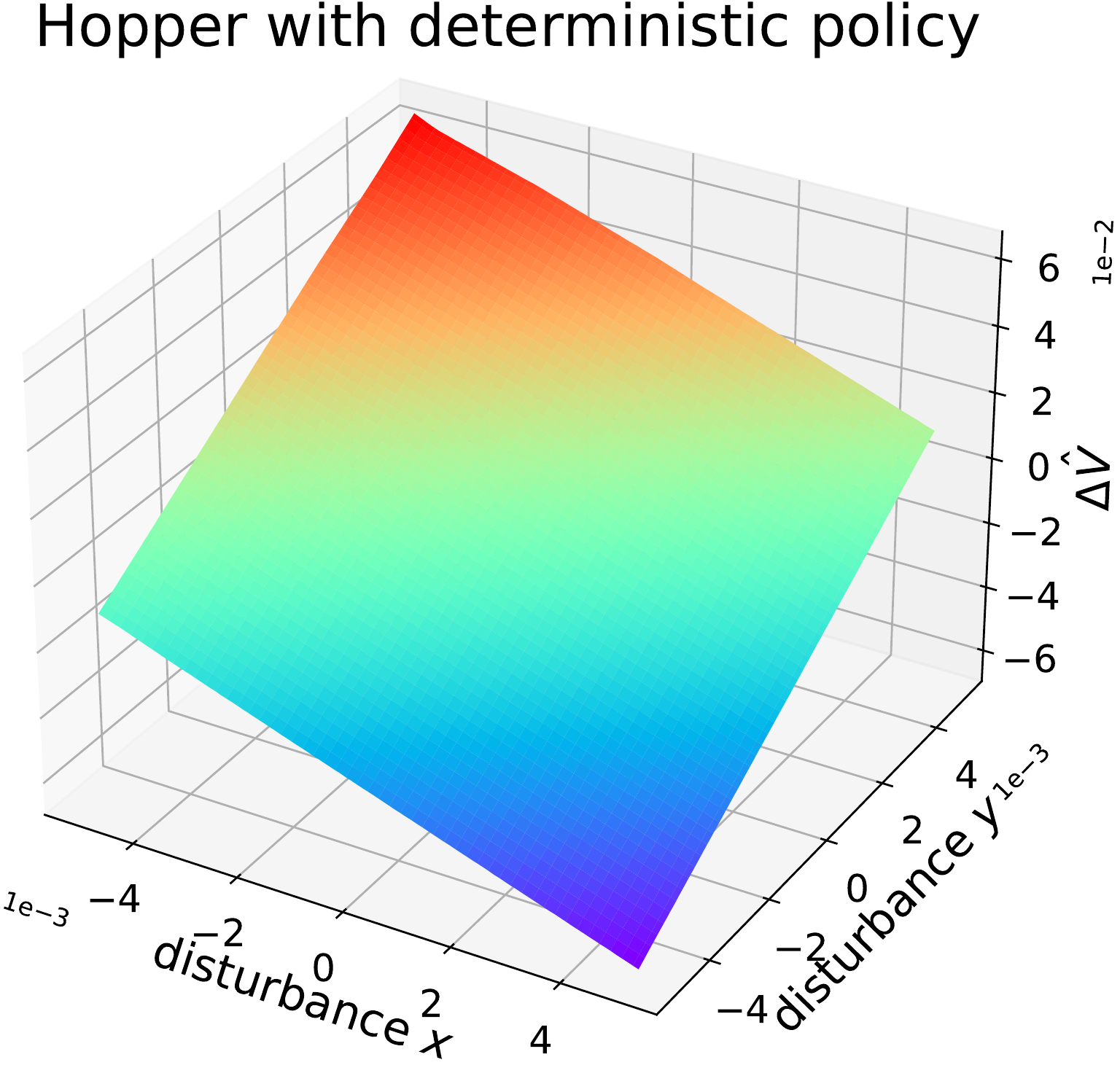}}
    \subfigure[Stochastic policy.]{
        \label{fig: stochastic}
        \includegraphics[height=0.2\textwidth]{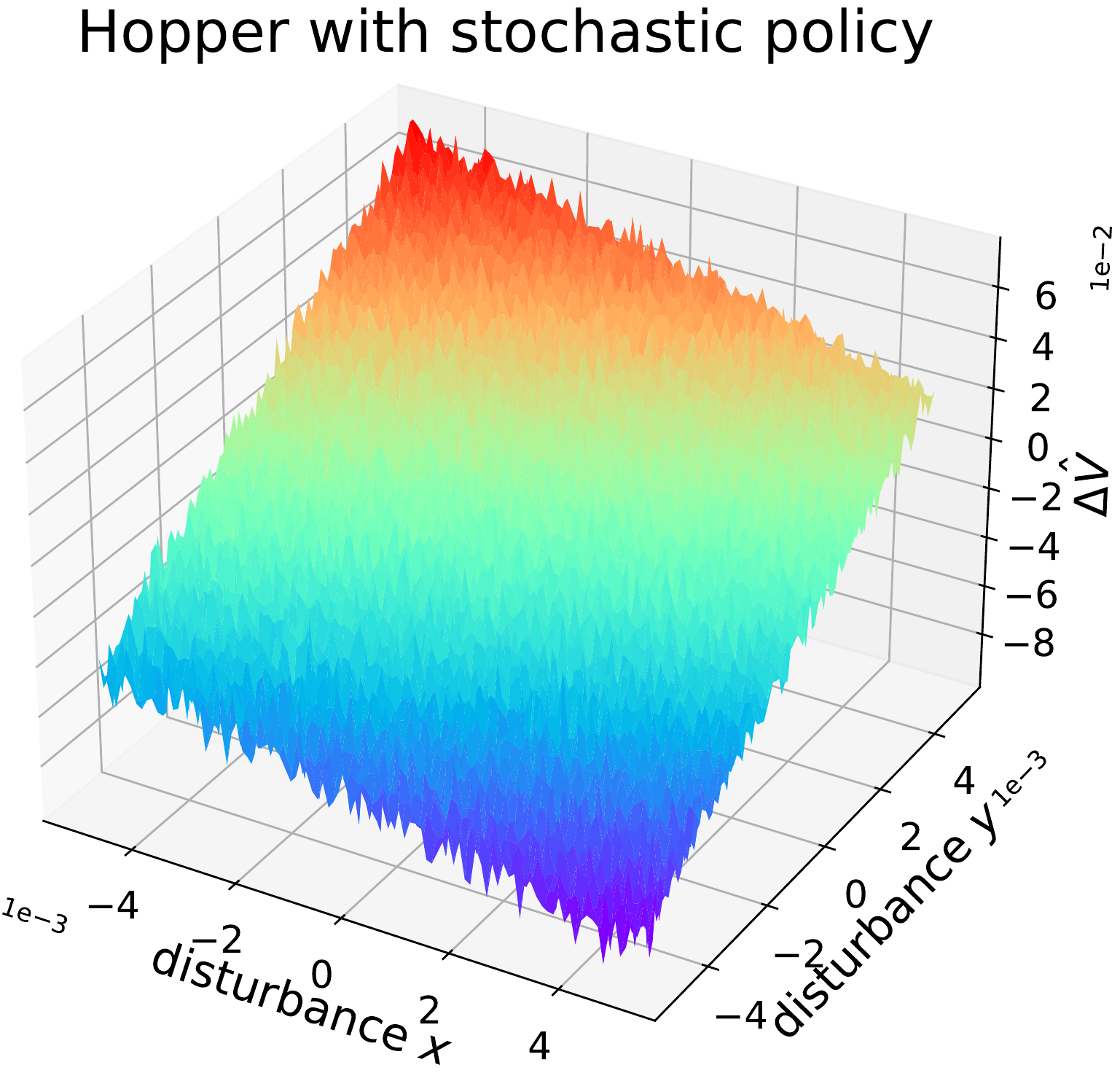}
    }
\caption{Visualize the  local linear property of critics in SAC (notations refer to Section \ref{sec:sc_sac}). We randomly choose a state $s$ and two mutually perpendicular directions $x,y$ in the normalized state space, and visualize the value change $\Delta \hat{V}\triangleq \mathbb{E}_{\delta}\left[\Delta Q_\theta^{\min} \left(s^\prime,f_\phi\left(\delta;s\right)\right)\right]$ in $\text{span}(x,y)$. In Figure \ref{fig: deterministic}, we make the trained policy in SAC to be deterministic (i.e., let $\delta=0$). In Figure \ref{fig: stochastic}, we sample $\delta\sim \mathcal{N}(0,1)$ 1000 times to estimate the expectation for each point.}
\label{fig: linear}
\end{figure}

\textbf{Does the Approximation Make Sense in Practice?} In our experiments, we find that the local linear property of ${U}_{\theta,\phi}(s)$ for small $\epsilon$ is well satisfied for existing actor-critic algorithms (we use the popular off-policy algorithm SAC as an instance to illustrate in Figure \ref{fig: linear}), and this approximation achieves satisfactory performance as reported in Section \ref{sec: experiments}.

\setcounter{algorithm}{2}
\begin{algorithm}[htb]
	\caption{State-Conservative Soft Actor-Critic}
	\label{alg: sc-sac} 
	\begin{algorithmic}[1]
	\STATE \textbf{Input:} 
     Critic $Q_{\theta_1},Q_{\theta_2}$. Actor $\pi_\phi$. Initial temperature parameter $\alpha$. Step size $\beta_Q, \beta_\pi, \beta_\alpha$. Target smoothing coefficient $\tau$. 
		\STATE $\bar{\theta}_1\leftarrow\theta_1, \bar{\theta}_2\leftarrow\theta_2,
		\mathcal{D}\leftarrow\emptyset$.
		\FOR{each iteration}
		\FOR{each environment step}
		\STATE $ {a}_t \sim \pi(\cdot| {s}_t)$, ${s}_{t+1} \sim p_0(\cdot| {s}_t,  {a}_t)$.
		\STATE $\mathcal{D}\leftarrow \mathcal{D}\cup\{ {s}_t, {a}_t, r({s}_t, {a}_t), {s}_{t+1}\}$.
		\ENDFOR
		\FOR{each training step}
		\STATE $\theta_i \leftarrow \theta_i - \beta_Q {\nabla}J_{\epsilon\text{-}Q}(\theta)$ for $i=1,2$. 
		\STATE $\phi \leftarrow \phi - \beta_\pi {\nabla}J_{\epsilon\text{-}\pi}(\phi)$.
		\STATE $\alpha \leftarrow \alpha - \beta_\alpha {\nabla}J_\alpha(\alpha)$.
		\STATE $\bar{\theta}_i \leftarrow \tau\theta_i + (1-\tau)\bar{\theta}_i$ for $i=1,2$.
		\ENDFOR
		\ENDFOR
	\STATE \textbf{Output:} 
	$\theta_1,\theta_2,\phi$ 
	\end{algorithmic}
\end{algorithm}

\subsection{State-Conservative SAC: An Instance}
\label{sec:sc_sac}

Now we apply SCPO proposed in Section \ref{sec:scpo} to existing actor-critic algorithms to learn robust policies. Specifically, we present the SCPO-based algorithm \textbf{S}tate-\textbf{C}onservative \textbf{S}oft \textbf{A}ctor-\textbf{C}ritic (SC-SAC) as an instance \cite{sac}, where we use the GBR proposed in Section \ref{sec: GBI} for seek of efficient implementation.

Soft Actor-Critic (SAC) \cite{sac} is an off-policy actor-critic algorithm based on the maximum entropy RL framework,
where the actor aims to maximize both the expected return and the entropy. 
According to SAC, we reparameterize the policy as ${a}_{t}= f_{\phi}\left(\delta_{t} ; {s}_{t}\right)$, where $\delta_t$ is standard Gaussian noise; 
we use the target Q-value network $Q_{\bar{\theta}}$ whose parameters $\bar{\theta}$ is updated in a moving average
fashion; 
we use the automatic tuning for the temperature parameter $\alpha$ and the clipped double Q-learning $Q_\theta^{\min}(s,a)\triangleq\min_{i=1,2} Q_{\theta_i}(s,a)$ \cite{td3}.

In the policy evaluation step, we train the critic by minimizing the Bellman residual $J_{\epsilon\text{-}Q}$ defined in Equation (\ref{equ:critic}), with the target $\widehat{V}(s_{t+1})$ in Equation (\ref{equ: target-v})  now defined as
\begin{align}\label{equ: sc-soft-v}
    \widehat{V}(s_{t+1}) \triangleq \  \mathbb{E}_{\delta_t\sim\mathcal{N}}\left [Q_{\bar{\theta}}^{\text{min}}\left(s_{t+1},f_\phi\left(\delta_{t+1};s_{t+1}\right)\right) \right. 
    -\epsilon\nabla_{s} Q_{\bar{\theta}}^{\text{min}}\left(s_{t+1},f_\phi\left(\delta_{t+1};s_{t+1}\right)\right)
    \left. -  \alpha\log\pi_\phi\left(f_\phi\left(\delta_{t+1};s_{t+1}\right)|s_{t+1}\right)\right].
\end{align}
In the policy improvement step, we train the actor by maximizing the state-conservative entropy regularized objective
    \begin{align}\label{equ: sc-actor}
    J_{\epsilon\text{-}\pi}(\phi)\triangleq\mathbb{E}_{s_t\sim D,\delta_t\sim \mathcal{N}}\left[Q_{{\theta}}^{\text{min}}\left(s_t,f_\phi\left(\delta_t;s_t\right)\right) \right. 
    -\left. \epsilon\nabla_s Q_{{\theta}}^{\text{min}}\left(s_t,f_\phi\left(\delta_t;s_t\right)\right) -  \alpha\log\pi_\phi\left(f_\phi\left(\delta_t;s_t\right)|s_t\right)\right] 
    \end{align}
In the temperature $\alpha$ update step, we tune $\alpha$ by minimizing 
\begin{align}\label{obj:alpha}
J_\alpha(\alpha) \triangleq -\alpha\,\mathbb{E}_{s_t\sim\mathcal{D}, \delta_t\sim \mathcal{N}}\left[ \log\pi_\phi\left(f_\phi\left(\delta_t;s_t\right)|s_t\right) + \mathcal{H} \right],
\end{align}
where $\mathcal{H}$ is the target value of the entropy term. We show the pseudo code of SC-SAC in Algorithm \ref{alg: sc-sac}.

\begin{figure*}[htb]
\centering
\includegraphics[width=0.76\textwidth]{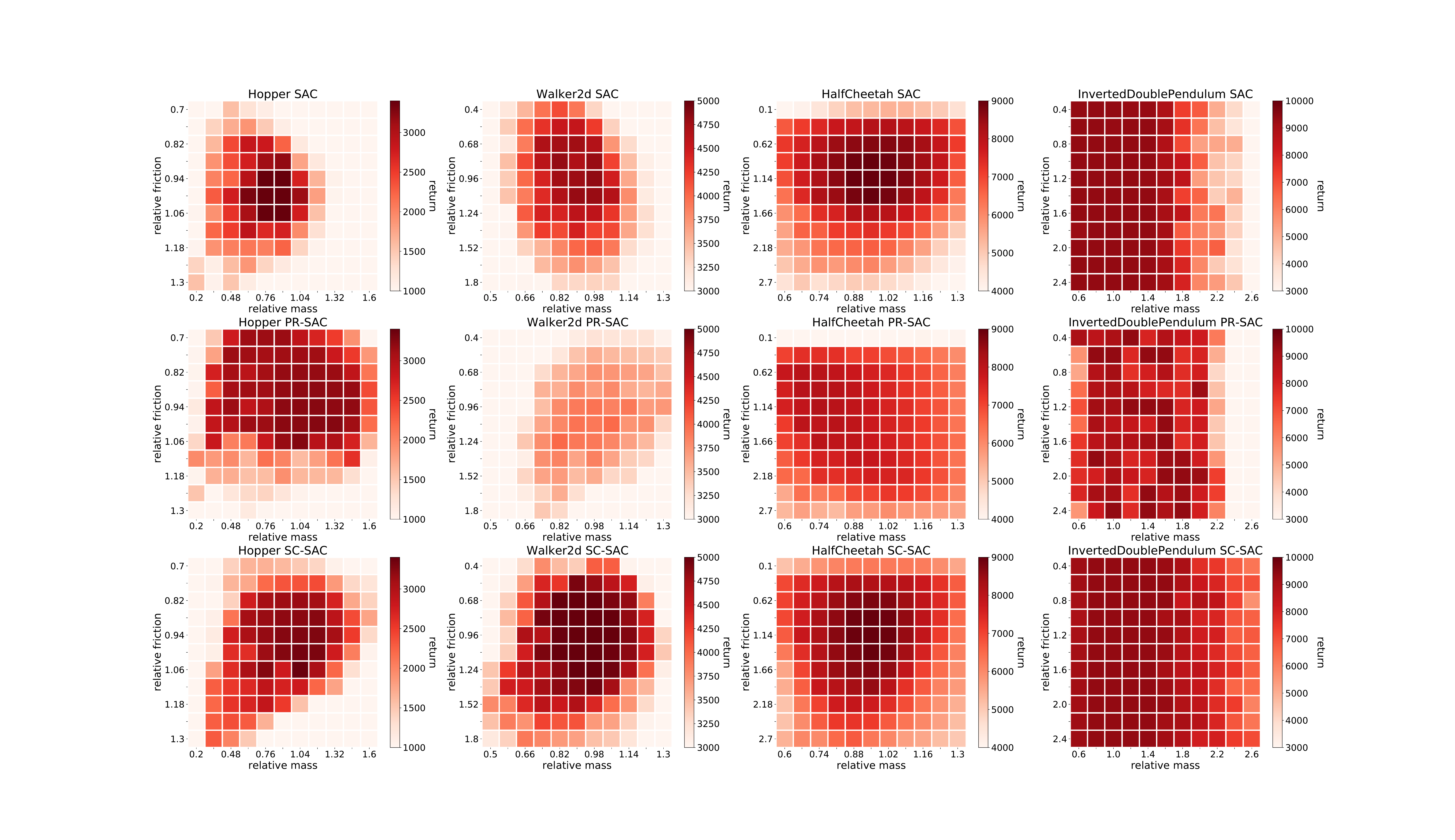}
\caption{
Compare SAC, PR-SAC and SC-SAC in target environments with perturbed parameters (mass, friction). 
We randomly choose $4$ policies from the last $10$ epochs for each seed and run $8$ episodes for each policy. 
That is, each point here is evaluated over $160$ episodes ($5\times4\times8$). 
More details for the implementations of PR-SAC can be found in Appendix C. 
The results show that SC-SAC trained policies are more robust than the original SAC and  the PR-SAC in perturbed target environments. 
}
\label{fig: heatmap}
\end{figure*}

\begin{figure}[htb]
\centering
\includegraphics[width=0.47\textwidth]{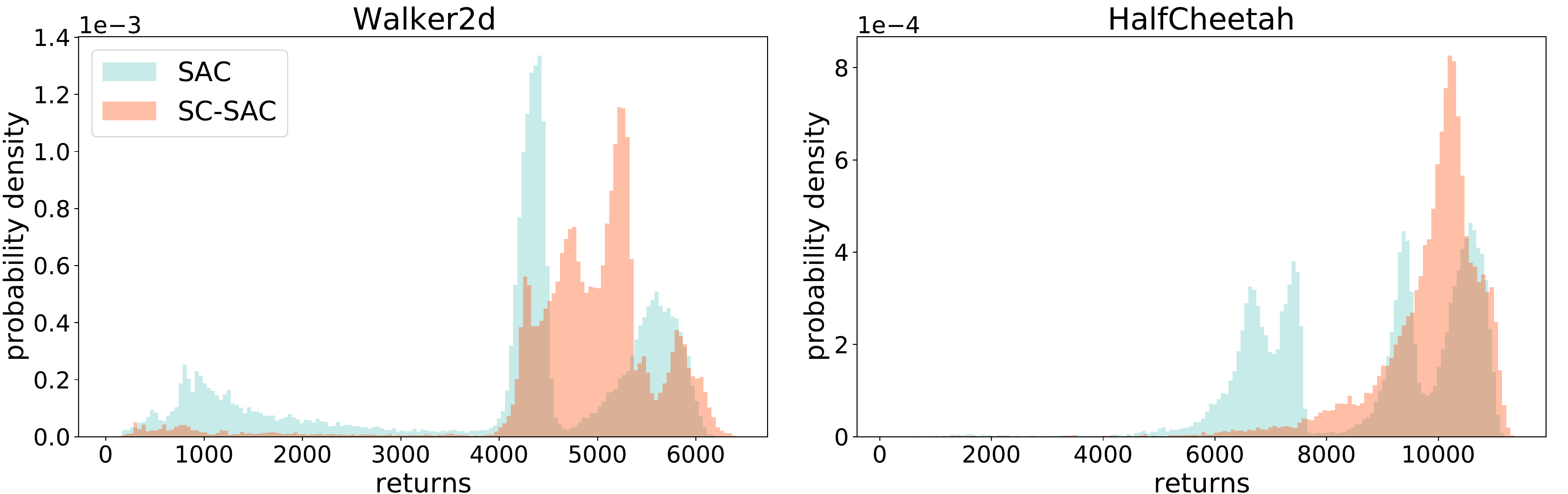}
\caption{Compare the return distribution between SAC and SC-SAC with truncated Gaussian mass and friction. 
We evaluate the return $10^5$ times for each task with policy sampled from last $10$ epochs and environmental parameters sampled from truncated Gaussian distributions (see Appendix C for details). 
The results show that the return distribution of SC-SAC is more concentrated in the high-value areas.
}
\label{fig: distribution}
\end{figure}

\begin{figure}[htb]
\centering
\includegraphics[width=0.47\textwidth]{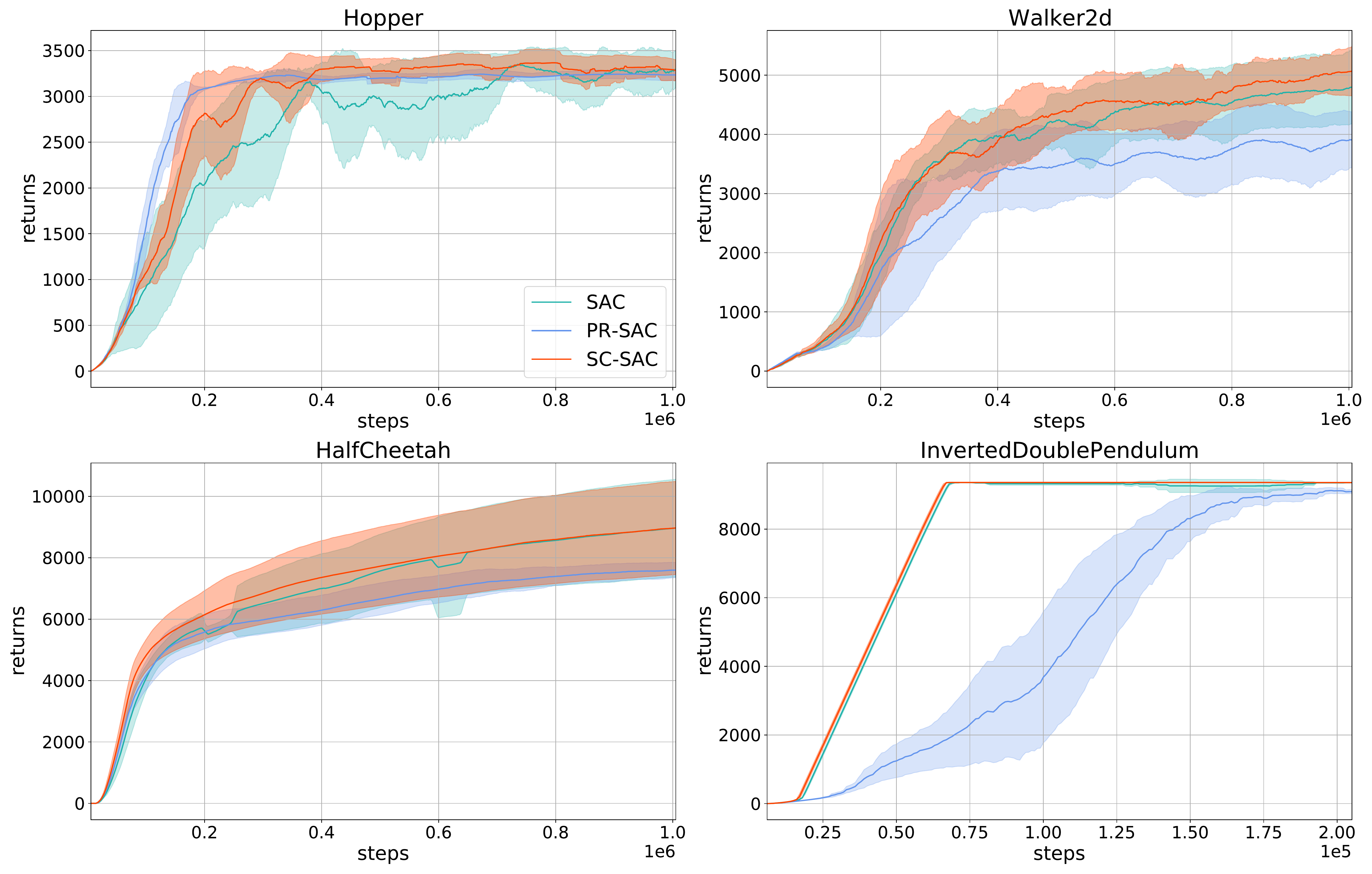}
\caption{Training curves for different algorithms. The results show that SC-SAC achieves almost identical performance as SAC in source environments for all tasks, while PR-SAC leads to performance degradation for several tasks.}
\label{fig: performance}
\end{figure}

\begin{figure}[htb]
\centering
\includegraphics[width=0.47\textwidth]{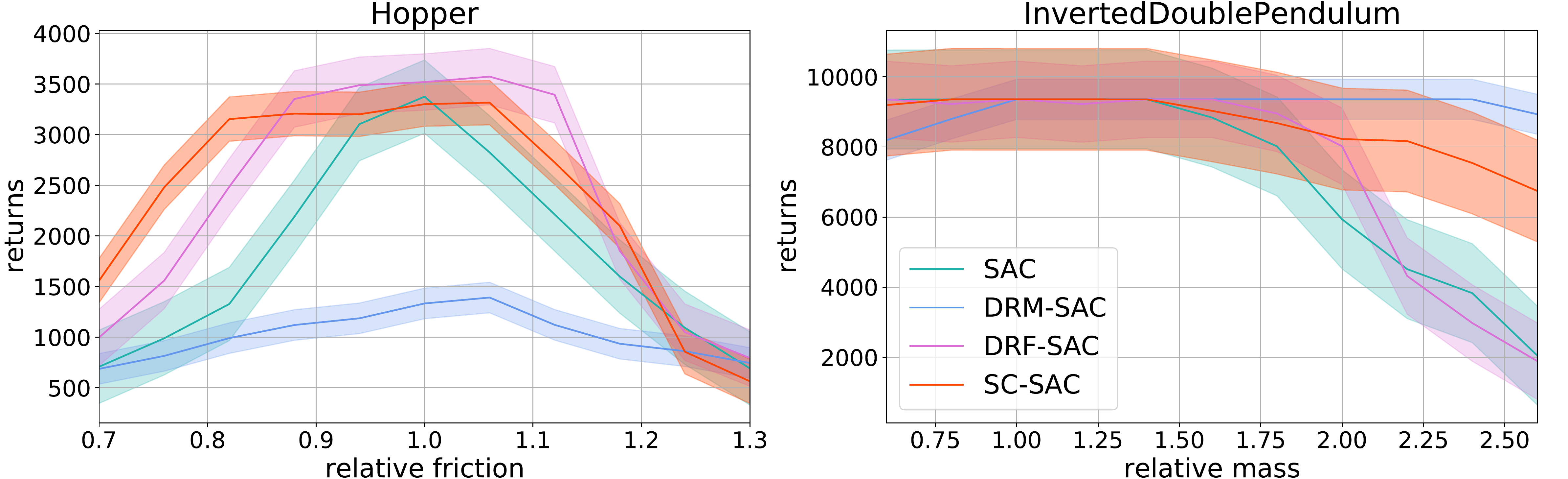}
\caption{Compare SCPO with DR in target environments with one parameter perturbed. DR uniformly randomizes the mass (DRM-SAC) or the friction (DRF-SAC) in the range given in Figure \ref{fig: heatmap} during training in source environments, while SC-SAC only uses a gradient based regularizer. We then evaluate them in target environments with perturbed mass or friction. The results show that SCPO learned policies are more robust to unmodeled disturbance than DR.}
\label{fig: domain randomization}
\end{figure}

\section{Experiments}\label{sec: experiments}
In this section, we conduct experiments on the SCPO-based algorithm SC-SAC in several MuJoCo benchmarks \cite{mujoco} to evaluate its performance.

\textbf{Hyperparameter Setting} 
The hyperparameter $\epsilon$ serves as a regularization coefficient in SCPO. 
By the definition in Section \ref{subsec:state_conservative_objective}, larger $\epsilon$ implies higher intensity of the disturbance considered in SC-MDP. 
However, too large $\epsilon$ can lead to suboptimal policies and thus degraded performance. Thus, we tune the hyperparameter $\epsilon$ in the Hopper-v2 task by grid search and find that it achieves the best performance when $\epsilon=0.005$. 
We then \textit{set $\epsilon=0.005$ for all the tasks in our experiments}. 
See Section \ref{sec: ablation} for sensitivity analysis.

\textbf{Implementation and Evaluation Settings} 
We normalize the observations for both SAC and SC-SAC in all tasks. We keep all the parameters in SC-SAC the same as those in original SAC. We train policies for $200$k steps (i.e., $200$ epochs) in InvertedDoublePendulum-v2 and $1000$k steps (i.e., $1000$ epochs) in other tasks. 
We train policy for each task with 5 random seeds. 
We perturb the parameters (e.g., mass and friction) in target environments to generate disturbance in transition dynamics.
More details for implementation and evaluation settings can be found in Appendix C.

\subsection{Comparative Evaluation}\label{sec: comparative}

We compare SC-SAC with the original SAC and PR-MDP \cite{tessler2019action} in this part.

\textbf{Comparison with Original SAC} 
We compare SC-SAC with original SAC in target environments with perturbed environmental parameters (mass, friction).
Figure \ref{fig: heatmap} illustrates that \textit{SC-SAC is more robust than SAC when generalize to target environments}. 
We then evaluate policies in target environments with truncated Gaussian distributed mass and friction, since the perturbation on environmental parameters in the real world usually follows Gaussian distribution. Results in Figure \ref{fig: distribution} demonstrate that the return distributions of SC-SAC are more concentrated in high-value areas than that of SAC.
Note that \textit{SC-SAC does not use any prior knowledge} about the disturbance (e.g., mass and friction changes) during training. We further compare the computational efficiency between SC-SAC and SAC in Appendix D.1.

\textbf{Comparison with the Action Robust Algorithm} 
Action robust (AR) algorithms learn robust policies by using the disturbance in action space to approximate the unmodeled disturbance in transition dynamics. 
We apply the state-of-the-art AR algorithm PR-MDP, which uses an adversarial policy to disturb actions during training, to SAC (called PR-SAC) and compare it with SC-SAC in Figure \ref{fig: heatmap}. 
Results show that \textit{SC-SAC achieves higher average returns than PR-SAC in most tasks}. This is mainly because that adversarial training tends to result in degraded performance, while regularizer-based SCPO maintains the original performance of SAC more easily (see Figure \ref{fig: performance} for the training curves).

\textbf{Comparison with Domain Randomization}
The domain randomization (DR) algorithm randomizes parameters in source environments during training to learn policies that are robust to perturbations on these parameters. 
However, when testing in target environments, we can suffer from unmodeled disturbance that comes from perturbations on unseen parameters. 
To compare the robustness against unmodeled disturbance between SCPO and DR, we train SAC with one parameter (mass or friction) uniformly randomized in the range given in Figure \ref{fig: heatmap} and then test it in target environments with the other parameter perturbed. 
As shown in Figure \ref{fig: domain randomization}, DR trained policies are robust to the disturbance that is modeled in advance but perform poorly when generalize to unmodeled disturbance. By contrast, \textit{though the perturbations on both mass and friction are unmodeled during training, SC-SAC trained policies are robust to them consistently}.

\begin{figure}[htb]
\centering
\includegraphics[width=0.47\textwidth]{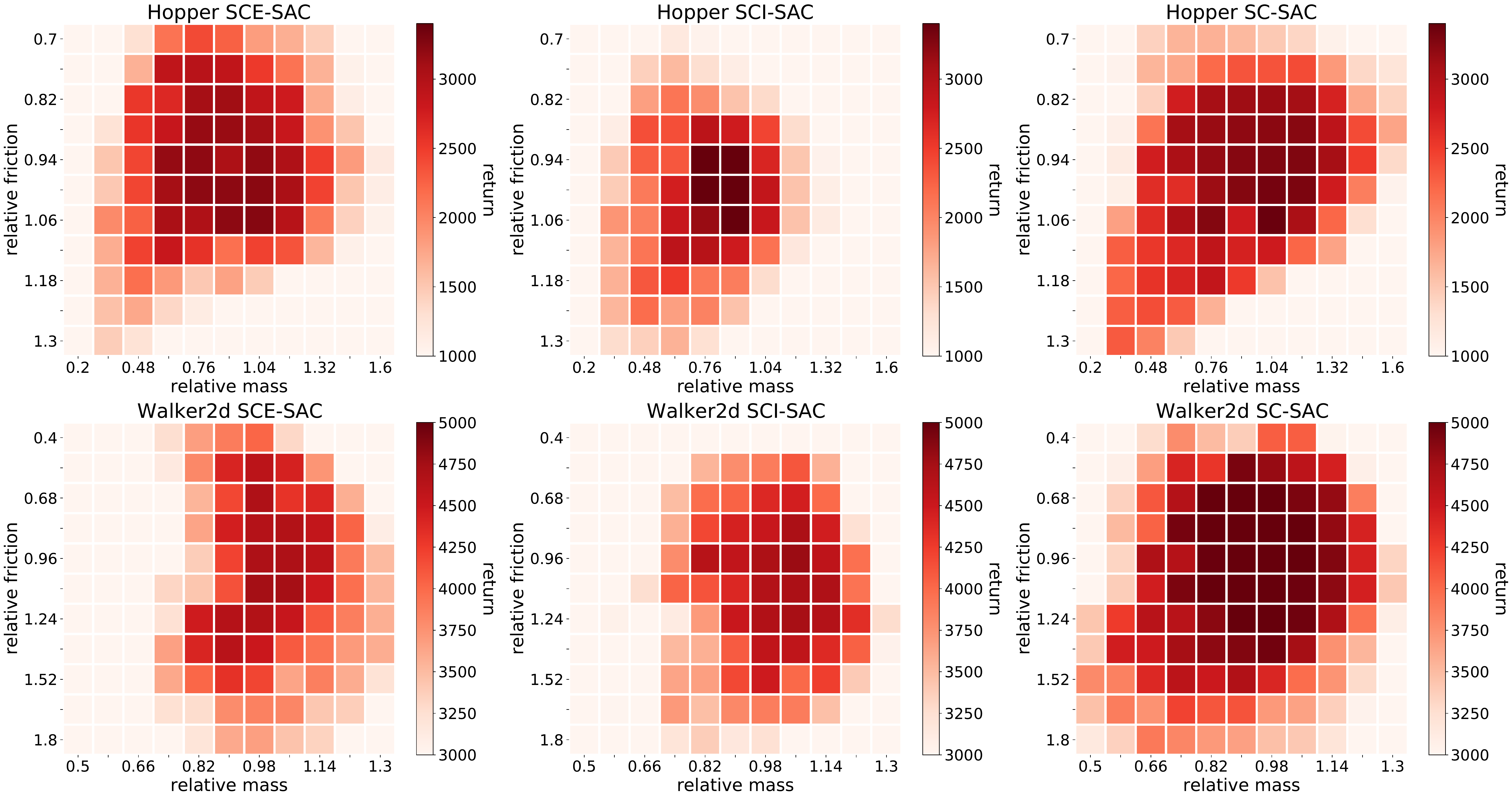}
\caption{Evaluate polices trained with only the state-conservative policy evaluation step (SCE-SAC) or the  state-conservative policy improvement step (SCI-SAC). 
Each point here is evaluated over $160$ episodes (similar to that in Figure \ref{fig: heatmap}). 
Results show that both the SCE step and the SCI step in SCPO help to learn robust policies during training. 
}
\label{fig: eval_impr}
\end{figure}

\begin{figure}[htb]
\centering
\includegraphics[width=0.47\textwidth]{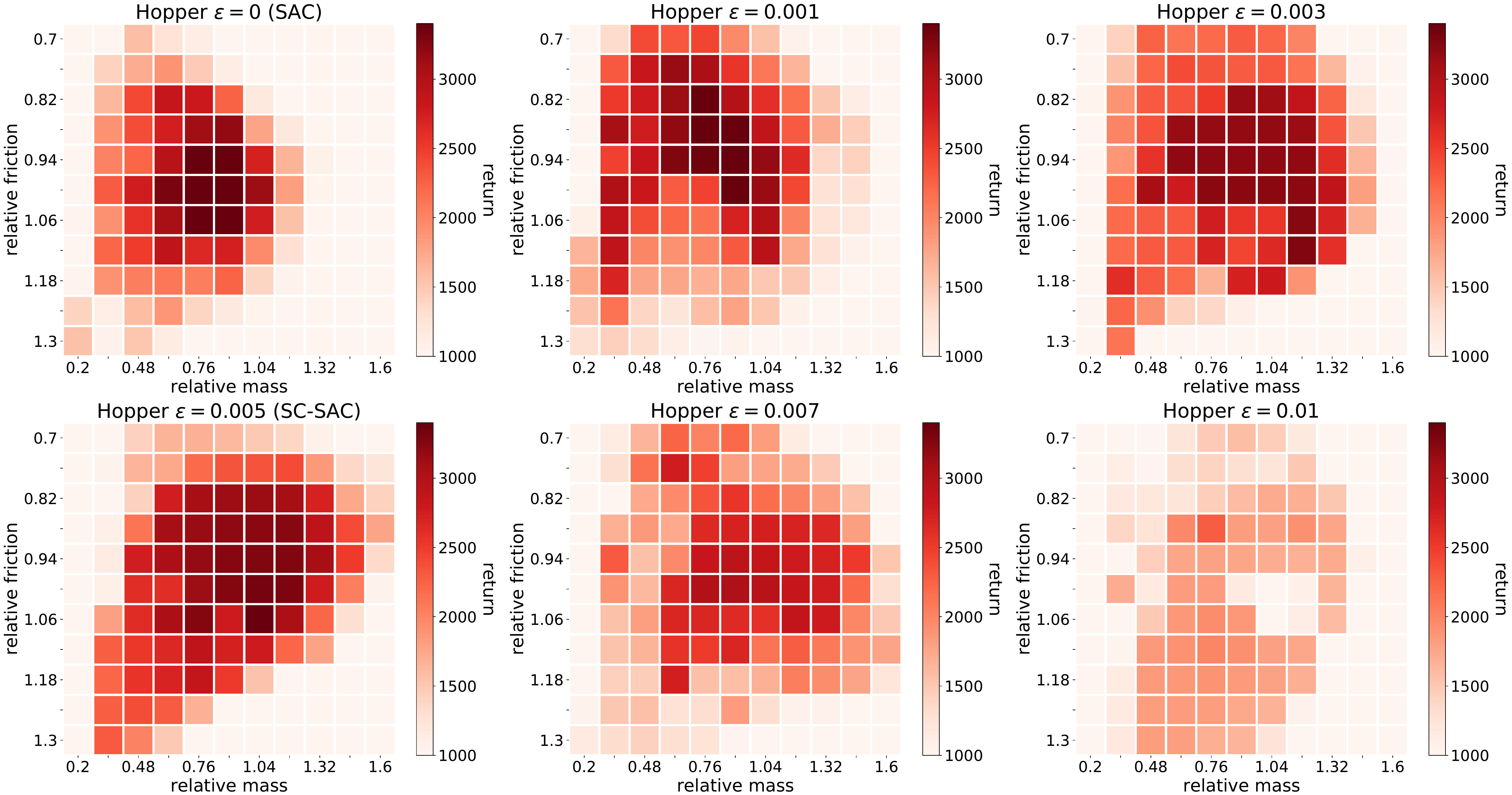}
\caption{Sensitivity analysis on the hyperparameter $\epsilon$. We choose $\epsilon=0.001, 0.003, 0.005, 0.007, 0.01$ in SC-SAC and test in perturbed target environments. Each point is evaluated over $160$ episodes. 
Results show that SC-SAC is relatively insensitive to $\epsilon$ within a specific range, while a too large $\epsilon$ hurts the performance of SAC in source environments.
}
\label{fig: ablation}
\end{figure}

\subsection{Ablation Study} \label{sec: ablation}

In this part, we conduct ablation study on different components of SC-SAC to analyze their effects empirically.

\textbf{Policy Evaluation v.s. Policy Improvement} SCPO based algorithms include a state-conservative policy evaluation step and a state-conservative policy improvement step. We conduct ablation study to analyze their effects in SCPO respectively.
Specifically, we apply SCPO to SAC with only the policy evaluation step (SCE-SAC) or the policy improvement step (SCI-SAC) and then report their performance in perturbed target environments in Figure \ref{fig: eval_impr}. 
Results show that \textit{both the policy evaluation step and the policy improvement step in SCPO play an important role in the training process}.

\textbf{Sensitivity Analysis on $\epsilon$} 
We analyze the sensitivity of SC-SAC to the hyperparameter $\epsilon$. We evaluate policies trained with different $\epsilon$ in perturbed environments and show the results in Figure \ref{fig: ablation}. 
Results demonstrate that \textit{SC-SAC is relatively insensitive to $\epsilon$ within a specific range ($\epsilon\in [0.001, 0.005]$)}, while \textit{with too large epsilon ($\epsilon\geq 0.01$) SCPO can result in suboptimal policies} that perform poorly in target environments. In practice, we find an $\epsilon$ that makes the value of GBR be around $0.5\%$ of the original objective can usually serve as a good initial value for grid search.

\section{Conclusion}

In this paper, we propose the state-conservative MDP (SC-MDP) to learn robust policies against disturbance in transition dynamics without modeling the disturbance in advance. 
To solve continuous control tasks, we further propose the State-Conservative Policy Optimization (SCPO) algorithm and approximately solve it via a simple Gradient Based Regularizer (GBR). Experiments show that the SCPO learns robust policies against the disturbance in transition dynamics. 
Applying SCPO to on-policy and offline reinforcement learning algorithms is exciting avenues for future work.

\section*{Acknowledgements}

We would like to thank all the anonymous reviewers for their insightful comments. This work was supported in part by National Science Foundations of China grants U19B2026, U19B2044, 61822604, 61836006, and 62021001, and the Fundamental Research Funds for the Central Universities grant WK3490000004.

\bibliography{aaai22}
\bibliographystyle{aaai22}

\newpage
\appendix

\section{Proofs for Propositions}\label{sec:proof}

\subsection{Derivation from Equation (\ref{equ:emp_wass_bell}) to Equation (\ref{equ:emp_wass_bell_dual})}
\begin{proof}
    According to the duality theory in convex optimization literature (see Chapter 5 of \citet{boyd2004convex}), we can consider the second term in Equation (6), i.e., 
    \begin{equation}\label{equ:convex_constrained_problem}
    \inf _{s^{\prime} \in B_{\epsilon}\left(s_{t+1}\right)} \mathbb{E}_{a^{\prime} \sim \pi\left(\cdot \mid s^{\prime}\right)}\left[Q\left(s^{\prime}, a^{\prime}\right)\right],
    \end{equation}
    as a constrained \emph{convex} optimization problem, since both the target function $\mathbb{E}_{a^{\prime} \sim \pi\left(\cdot \mid s^{\prime}\right)}\left[Q\left(s^{\prime}, a^{\prime}\right)\right]$ a convex function in $s^{\prime}$ and the constraint $B_{\epsilon}(s_{t+1})$ a convex region in $\mathcal{S}$ by our assumptions in Section 4.1. Now invoking the duality theory, we can conclude that the constraint convex optimization problem \eqref{equ:convex_constrained_problem} is equivalent to the following \emph{Lagrange dual problem},
    \begin{equation}
    \sup_{\lambda>0}\left\{\inf _{s^{\prime} \in \mathcal{S}} \mathbb{E}_{a^{\prime} \sim \pi\left(\cdot \mid s^{\prime}\right)}\left[Q\left(s^{\prime}, a^{\prime}\right)\right]+\lambda\left(d\left(s^{\prime}, s_{t+1}\right)^{p}-\epsilon\right)\right\},
    \end{equation}
    which is just the second term in Equation (5). This shows the equivalence between Equation (5) and Equation (6).
\end{proof}

\subsection{Proof of Proposition 2}
\begin{proof}
    By the definition of $\mathbb{T}^{\pi}_{\epsilon}$ in Equation ({8}), for any $Q_1,Q_2\in\mathbb{R}^{|\mathcal{S}\times\mathcal{A}|}$,  any $(s,a)\in\mathcal{S}\times\mathcal{A}$, we have
    \begin{equation}\label{equ:difference}
        \mathbb{T}^{\pi}_{\epsilon}Q_1(s,a)-\mathbb{T}^{\pi}_{\epsilon}Q_2(s,a)=
        \gamma \mathbb{E}_{\tilde{s}\sim p_0(\cdot|s,a)}
    \left[\inf_{s^{\prime}\in B_{\epsilon}(\tilde{s})}\mathbb{E}_{a^{\prime}\sim\pi(\cdot|s^{\prime})}[Q_1(s^{\prime},a^{\prime})]-\inf_{s^{\prime}\in B_{\epsilon}(\tilde{s})}\mathbb{E}_{a^{\prime}\sim\pi(\cdot|s^{\prime})}[Q_2(s^{\prime},a^{\prime})]\right].
    \end{equation}
    For any $\epsilon>0$ and any $\tilde{s}\in\mathcal{S}$, take ${s}^{\prime}_2(\tilde{s})\in B_{\epsilon}(\tilde{s})$ such that 
    \begin{equation}
        \inf_{s^{\prime}\in B_{\epsilon}(\tilde{s})}\mathbb{E}_{a^{\prime}\sim\pi(\cdot|s^{\prime})}[Q_2(s^{\prime},a^{\prime})]\geq\mathbb{E}_{a^{\prime}\sim\pi(\cdot|s_2^{\prime}(\tilde{s}))}[Q_2({s}^{\prime}_2(\tilde{s}),a^{\prime})]-\epsilon. 
    \end{equation}
    Taking this back to Equation (\ref{equ:difference}), we can derive that
    \begin{equation}
        \begin{aligned}
        \mathbb{T}^{\pi}_{\epsilon}Q_1(s,a)-\mathbb{T}^{\pi}_{\epsilon}Q_2(s,a)\leq& \gamma\mathbb{E}_{\tilde{s}\sim p_0(\cdot|s,a)}[\underbrace{\mathbb{E}_{a^{\prime}\sim\pi(\cdot|s_2^{\prime}(\tilde{s}))}[Q_1({s}^{\prime}_2(\tilde{s}),a^{\prime})]}_{{\text{definition of infimum}}}-\underbrace{\mathbb{E}_{a^{\prime}\sim\pi(\cdot|s_2^{\prime}(\tilde{s}))}[Q_2({s}^{\prime}_2(\tilde{s}),a^{\prime})]+\epsilon}_{ {\text{Equation (2)}}}]\\
     =&\gamma\mathbb{E}_{\tilde{s}\sim p_0(\cdot|s,a)}[\mathbb{E}_{a^{\prime}\sim\pi(\cdot|s_2^{\prime}(\tilde{s}))}[Q_1(s_2^{\prime}(\tilde{s}), a^{\prime})-Q_2(s_2^{\prime}(\tilde{s}), a^{\prime})]+\epsilon\\
        \leq&\|Q_1-Q_2\|_{\infty}+\epsilon.
        \end{aligned}
    \end{equation}
    Since $\epsilon>0$ is arbitrary, we conclude that 
    \begin{equation}\label{equ:1-2}
        \mathbb{T}^{\pi}_{\epsilon}Q_1(s,a)-\mathbb{T}^{\pi}_{\epsilon}Q_2(s,a)\leq \|Q_1-Q_2\|_{\infty},
    \end{equation}
    for any $(s,a)\in\mathcal{S}\times\mathcal{A}$. Similarly, we can derive that
    \begin{equation}\label{equ:2-1}
        \mathbb{T}^{\pi}_{\epsilon}Q_2(s,a)-\mathbb{T}^{\pi}_{\epsilon}Q_1(s,a)\leq \|Q_1-Q_2\|_{\infty},
    \end{equation}
    for any $(s,a)\in\mathcal{S}\times\mathcal{A}$. Thus, combining Equation (\ref{equ:1-2}) and (\ref{equ:2-1}), we get that
    $|\mathbb{T}^{\pi}_{\epsilon}Q_2(s,a)-\mathbb{T}^{\pi}_{\epsilon}Q_1(s,a)|\leq \|Q_1-Q_2\|_{\infty}$ for any $(s,a)\in\mathcal{S}\times\mathcal{A}$,
    which further implies that $\|\mathbb{T}^{\pi}_{\epsilon}Q_1-\mathbb{T}^{\pi}_{\epsilon}Q_1\|\leq \|Q_1-Q_2\|_{\infty}$. This completes the proof of Proposition 2.
\end{proof}
\subsection{Proof of Proposition 3}
\begin{proof}
    Recall that the state-conservative action value function satisfies the Bellman equation $Q^{\pi}_{\epsilon\text{-}\mathcal{S}}=\mathbb{T}^{\pi}_{\epsilon}Q^{\pi}_{\epsilon\text{-}\mathcal{S}}$.
    By iteratively applying the Bellman equation and using the relation between $\pi_{new}$ and $\pi_{old}$ given in Proposition 3, we have
    \begin{equation}
        \begin{aligned}
        Q^{\pi_{old}}_{\epsilon\text{-}\mathcal{S}}(\cdot|s_0,a_0)=&r(s_0,a_0)+\gamma\mathbb{E}_{s_1\sim p_0(s_0,a_0)}[\inf_{s_1^{\prime}\in B_{\epsilon}(s_1)}\mathbb{E}_{a^{\prime}_1\sim\pi_{old}(\cdot|s_1^{\prime})}[Q^{\pi_{old}}(s_1^{\prime},a_1^{\prime})]]\\
        \leq &r(s_0,a_0)+\gamma\mathbb{E}_{s_1\sim p_0(\cdot|s_0,a_0)}[\inf_{s_1^{\prime}\in B_{\epsilon}(s_1)}\mathbb{E}_{a_1^{\prime}\sim\pi_{new}(\cdot|s_1^{\prime})}[Q^{\pi_{old}}(s_1^{\prime},a_1^{\prime})]\\
        = & r(s_0,a_0)+\gamma\mathbb{E}_{s_1\sim p_0(\cdot|s_0,a_0)}[\inf_{s_1^{\prime}\in B_{\epsilon}(s_1)}\mathbb{E}_{a_1^{\prime}\sim\pi_{new}(\cdot|s_1^{\prime})}[r(s_1^{\prime},a_1^{\prime})\\
        &+\gamma\mathbb{E}_{s_2\sim p_0(\cdot|s_1^{\prime},a_1^{\prime})}[\inf_{s_2^{\prime}\in B_{\epsilon}(s_2)}\mathbb{E}_{a_2^{\prime}\sim \pi_{old}(\cdot|s_2^{\prime})}[Q^{\pi_{old}}(s_2^{\prime},a_2^{\prime})]]]]\\
        \leq &r(s_0,a_0)+\gamma\mathbb{E}_{s_1\sim p_0(\cdot|s_0,a_0)}[\inf_{s_1^{\prime}\in B_{\epsilon}(s_1)}\mathbb{E}_{a_1^{\prime}\sim\pi_{new}(\cdot|s_1^{\prime})}[r(s_1^{\prime},a_1^{\prime})\\
        &+\gamma\mathbb{E}_{s_2\sim p_0(\cdot|s_1^{\prime},a_1^{\prime})}[\inf_{s_2^{\prime}\in B_{\epsilon}(s_2)}\mathbb{E}_{a_2^{\prime}\sim \pi_{new}(\cdot|s_2^{\prime})}[Q^{\pi_{old}}(s_2^{\prime},a_2^{\prime})]]]]\\
        \leq &\cdots\,\,\cdots \,\,\,\,\,\,\,\,\,\,\,\, {\text{(By iteratively applying Bellman equation)}}\\
        \leq &Q^{\pi_{new}}_{\epsilon\text{-}\mathcal{S}}(s_0,a_0),
    \end{aligned}
    \end{equation}
    for any $(s_0,a_0)\in\mathcal{S}\times\mathcal{A}$. This completes the proof of Proposition 3.
\end{proof}

\section{Pseudo Codes}

\subsection{Discussions on the Relation between State and Transition Disturbance}

The equivalence holds in deterministic MDP as the Wasserstein distance simply reduces to the geometric distance in that case. However, in stochastic MDP it does not necessarily hold. For example, a stochastic transition that is close to a Dirac transition $\delta_s$ under Wasserstein distance can still have a small probability to go to states far away from $s$.

\subsection{Pseudo Code of SC-PI}
We conclude the pseudo code of the state-conservative policy iteration (SC-PI) algorithm in Algorithm \ref{alg:policy_iteration}.

\setcounter{algorithm}{0}
\begin{algorithm}[h]
    \caption{State-Conservative Policy Iteration}\label{alg:policy_iteration}
    \begin{algorithmic}[1]
    \STATE \textbf{Input:} A given MDP with transition $p_0$, an initial policy $\pi^0$, a non-negative real $\epsilon$ and iteration step $K$.
    \FOR{$k=0,1,\cdots,K$}
    \STATE Perform \textbf{State-Conservative Policy Evaluation} till convergence to obtain $Q^{\pi^k}_{\epsilon\text{-}\mathcal{S}}$.
    \STATE Perform \textbf{Policy Improvement} step by $\pi^{k+1}(\cdot|s)\leftarrow\argmax_{\pi\in\Delta_{\mathcal{A}}}\inf_{s^{\prime}\in B_{\epsilon}(s)}\mathbb{E}_{a^{\prime}\sim \pi}[Q^{\pi^k}_{\epsilon\text{-}\mathcal{S}}(s^{\prime},a^{\prime})]$.
    \ENDFOR
    \STATE \textbf{Output:} an estimation $\pi^{K+1}$ of the optimal policy.
    \end{algorithmic}
\end{algorithm}

\subsection{Pseudo Code of SCPO}

We conclude the pseudo code of the state-conservative policy optimization (SCPO) algorithm in Algorithm \ref{alg:  scpo}.

\begin{algorithm}[h]
	\caption{State-Conservative Policy Optimization}
	\label{alg: scpo} 
	 \begin{algorithmic}[1]
	    \STATE \textbf{Input:} Critic $Q_\theta$. Actor $\pi_\phi$. Step size $\beta_Q$, $\beta_\pi$.
		\STATE 
		$\mathcal{D}\leftarrow\emptyset$. 
		\FOR{each iteration}
		\FOR{each environment step}
		\STATE $ {a}_t \sim \pi(\cdot| {s}_t)$, ${s}_{t+1} \sim p_0(\cdot| {s}_t,  {a}_t)$.
		\STATE $\mathcal{D}\leftarrow \mathcal{D}\cup\{ {s}_t, {a}_t, r({s}_t, {a}_t), {s}_{t+1}\}$.
		\ENDFOR
		\FOR{each training step}
		\STATE $\theta \leftarrow \theta - \beta_Q {\nabla}J_{\epsilon\text{-}Q}(\theta)$.
		\STATE $\phi \leftarrow \phi - \beta_\pi {\nabla}J_{\epsilon\text{-}\pi}(\phi)$.
		\ENDFOR
		\ENDFOR
	\STATE \textbf{Output:} $\theta,\phi$.
	\end{algorithmic}
\end{algorithm}

\section{Details of Implementations}

\noindent\textbf{Observation Normalization}
We normalize the observations in SAC, PR-SAC and SC-SAC with fixed mean and standard deviation for each task. Specifically, We run SAC for $200$k steps in InvertedDoublePendulum-v2 and $1000$k steps in Hopper-v2, Walker2d-v2, and HalfCheetah-v2. We then use the training data in the replay buffer to calculate the mean and standard deviation in each dimension of observations. We normalize the observations by $o^\prime = \frac{o-\text{mean}}{\text{std}}$ in each dimension. 

\noindent\textbf{Implementation Details of PR-SAC} We apply PR-MDP to SAC in our experiments as it attains better performance in \cite{tessler2019action}. Note that we use a TD3-style \cite{td3} training process for the adversary, as directly optimizing the soft Q objective for the adversary only leads to a deterministic adversarial policy whose soft Q value diverges to negative infinity. We keep all hyperparameters the same as that in authors' implementation. That is, we set the probability of the adversary (the hyperparameter $\alpha$) to be $0.1$ and the training frequency of the adversary to be $10:1$.

\noindent\textbf{Shared Parameters} We keep all the shared parameters in SC-SAC and PR-SAC the same as those in original SAC \cite{sac}. We list these parameters and the corresponding values in Table \ref{hyperparameters} below.

\begin{table}[H]
	\caption{Shared parameters used in SAC, PR-SAC and SC-SAC.}
	\label{hyperparameters}
	\centering 
	\vspace{0.3cm}
	\begin{tabular}{ l  l }
		\toprule
		Parameter  & Value  \\ 
		\midrule 
		optimizer &Adam\\
		learning rate  & $ 3\cdot10^{-4} $\\
		discount ($ \gamma $)&0.99\\
		replay buffer size&$ 10^6 $\\
		number of hidden layers &2 \\
		number of hidden units per layer &256\\
		number of samples per minibatch &256\\
		entropy target & $-\dim(\mathcal{A})$\\
		nonlinearity & ReLU\\
		target smoothing coefficient &0.005\\
		target update interval & 1\\
		gradient steps & 1\\
		\bottomrule
	\end{tabular}
\end{table}

\begin{figure}[h]
\centering
\includegraphics[width=0.7\textwidth]{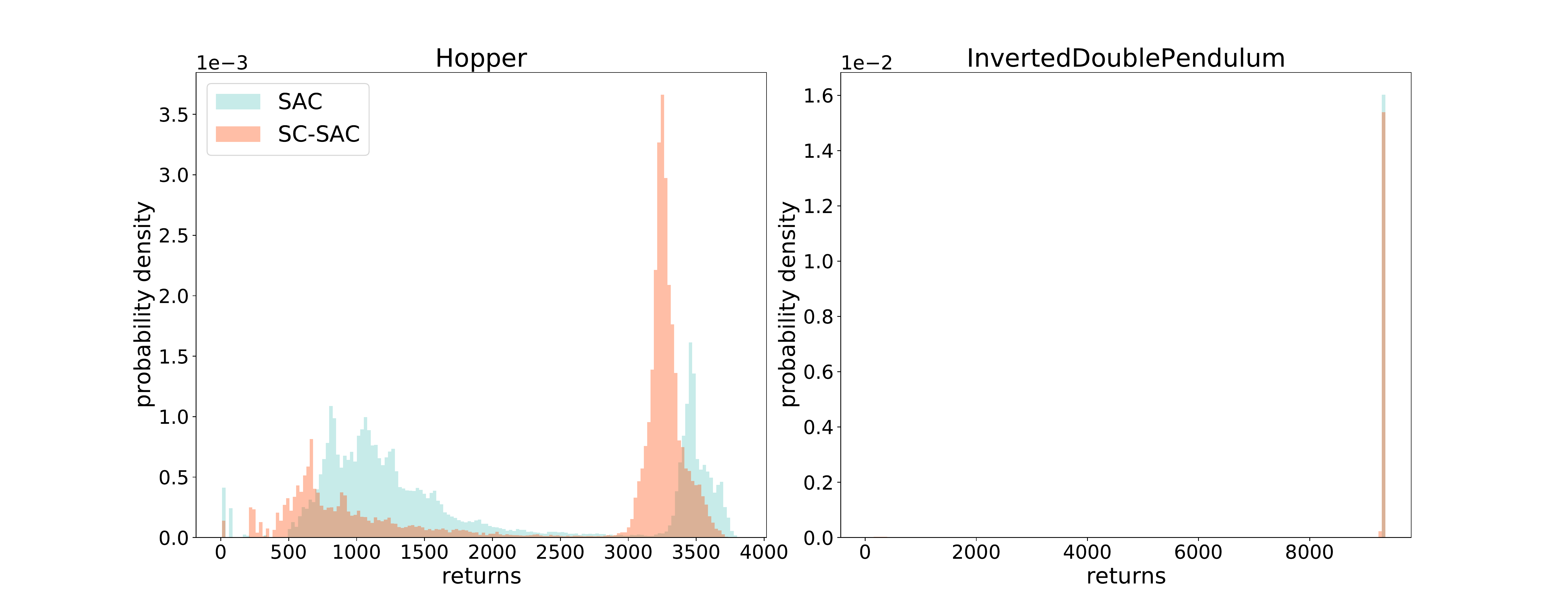}
\caption{Compare the return distribution between SAC and SC-SAC with truncated Gaussian mass and friction. 
}
\label{fig: distribution2}
\end{figure}

\begin{figure}[h]
\centering
\includegraphics[width=0.7\textwidth]{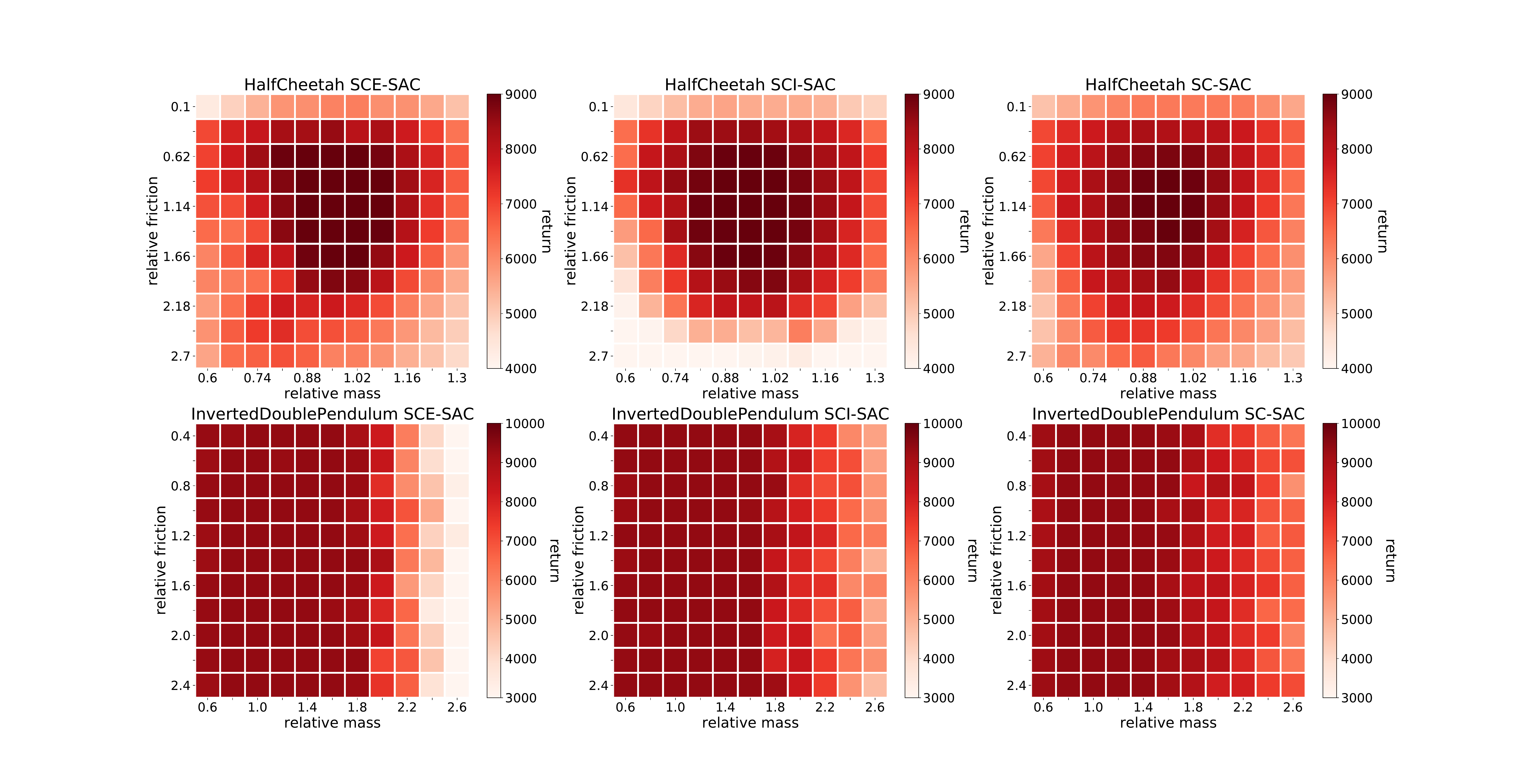}
\caption{Evaluate polices trained with only the state-conservative policy evaluation step (SCE-SAC) or the  state-conservative policy improvement step (SCI-SAC) in the HalfCheetah and the InvertedDoublePendulum tasks.
}
\label{fig: eval_impr_2}
\end{figure}

\begin{figure}[h]
    \centering
    \subfigure{
        \label{fig: hopper}
        \includegraphics[width=0.492\textwidth]{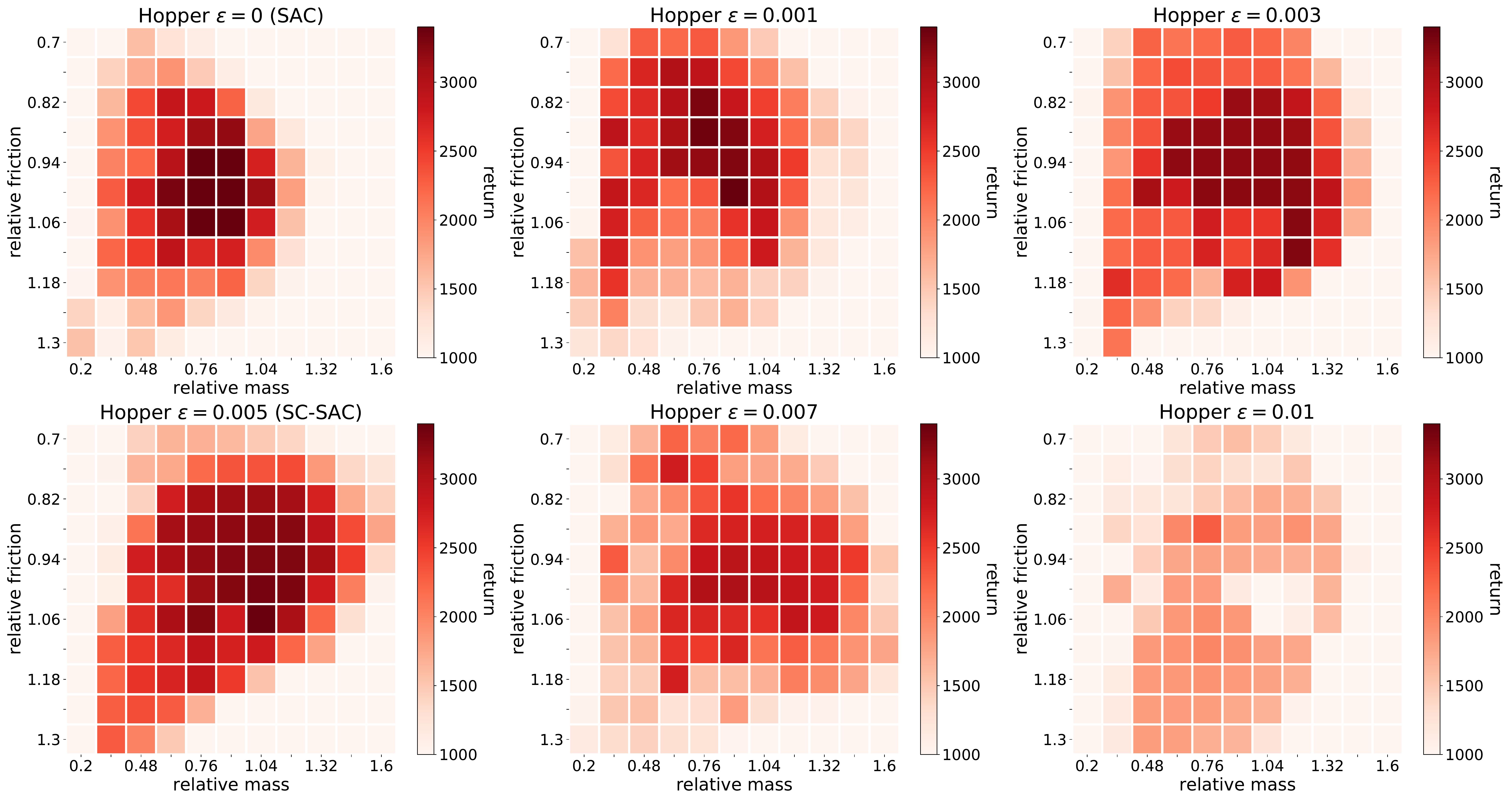}}
    \subfigure{
        \label{fig: walker}
        \includegraphics[width=0.492\textwidth]{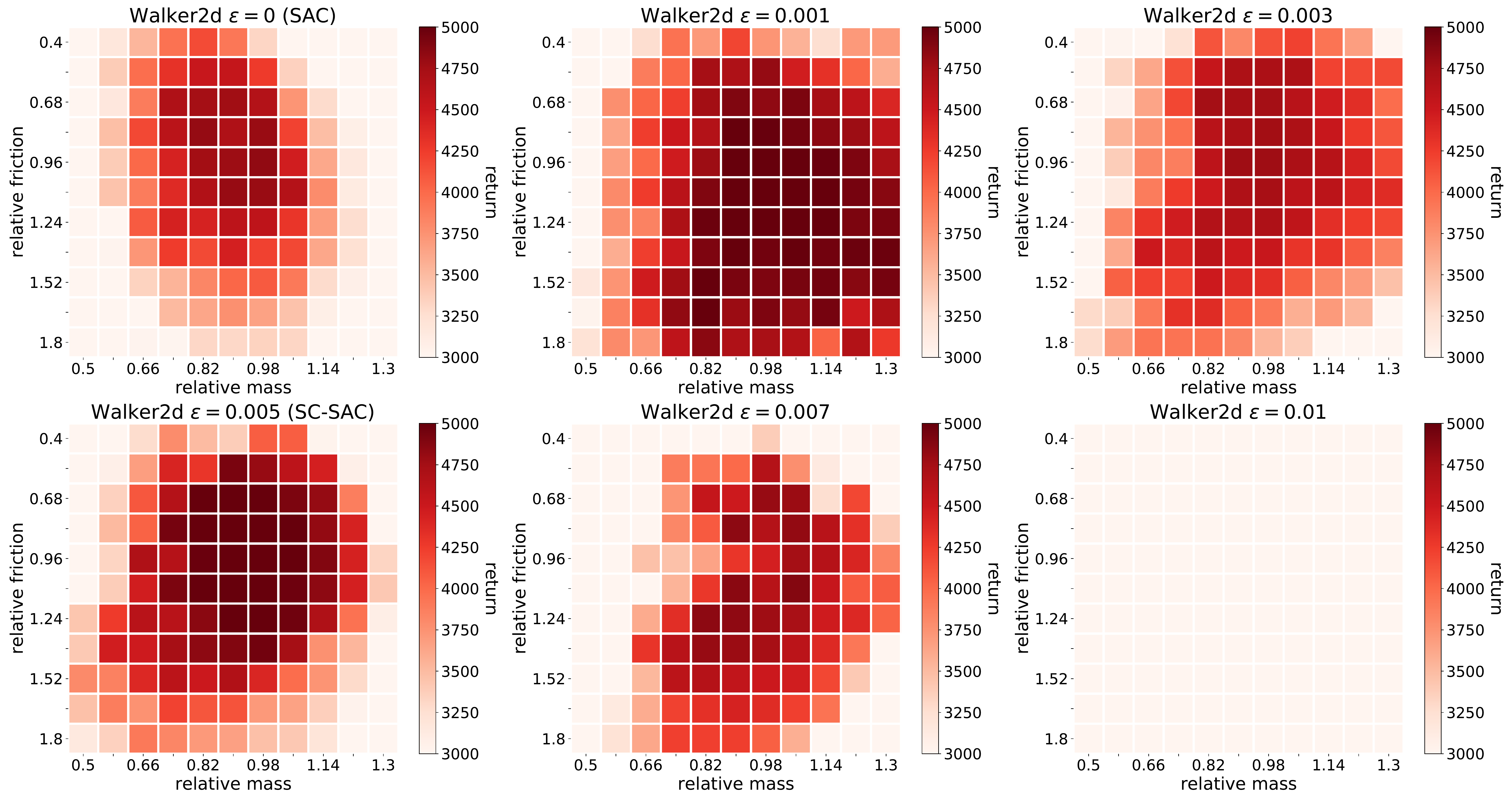}}
    \subfigure{
        \label{fig: cheetah}
        \includegraphics[width=0.492\textwidth]{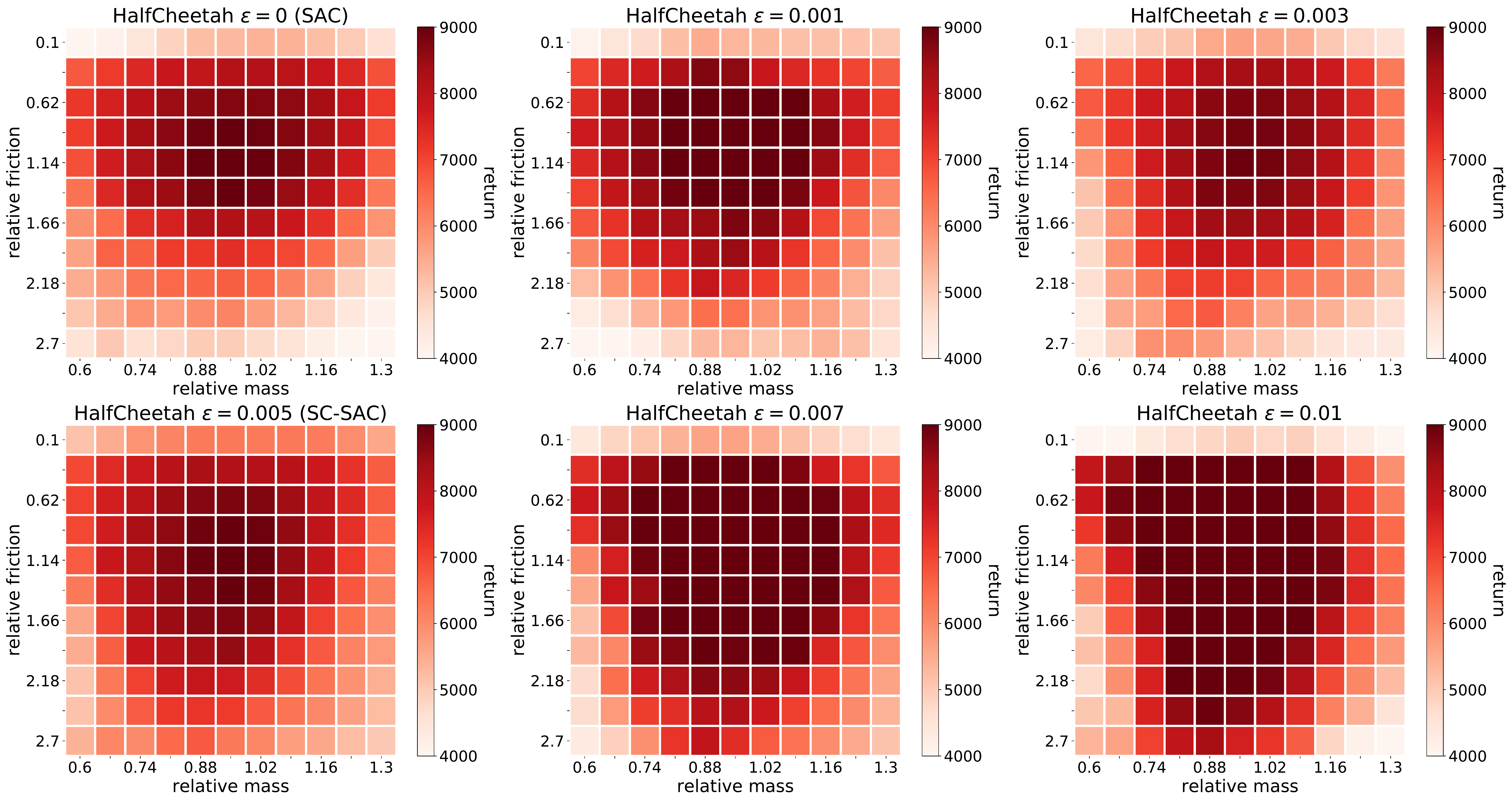}}
    \subfigure{
        \label{fig: pendulum}
        \includegraphics[width=0.492\textwidth]{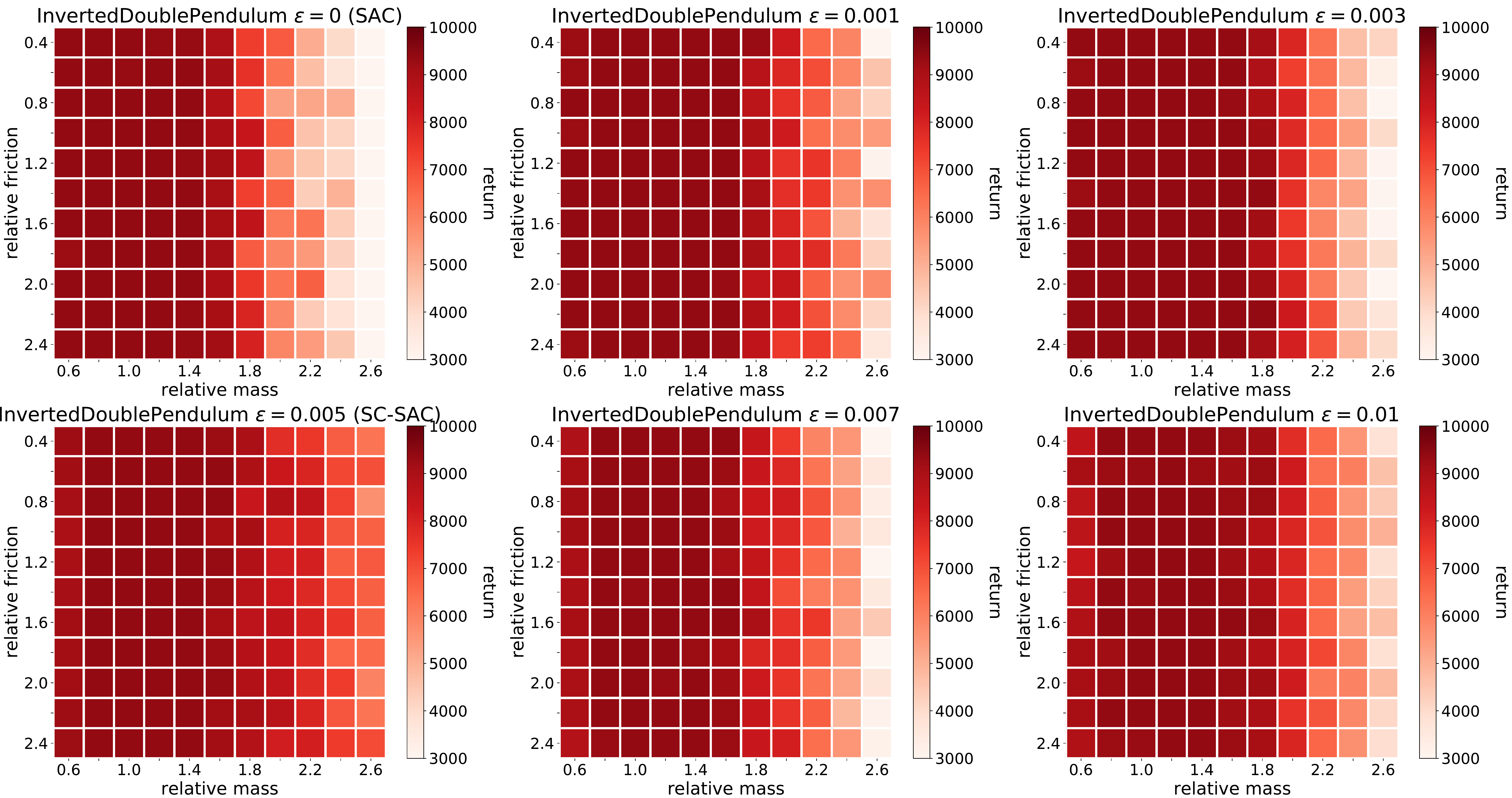}}
\caption{Sensitivity analysis on the hyperparamter $\epsilon$ in the HalfCheetah and the InvertedDoublePendulum tasks.}
\label{fig: ablation2}
\end{figure}

\begin{figure}[h]
\centering
\includegraphics[width=0.55\textwidth]{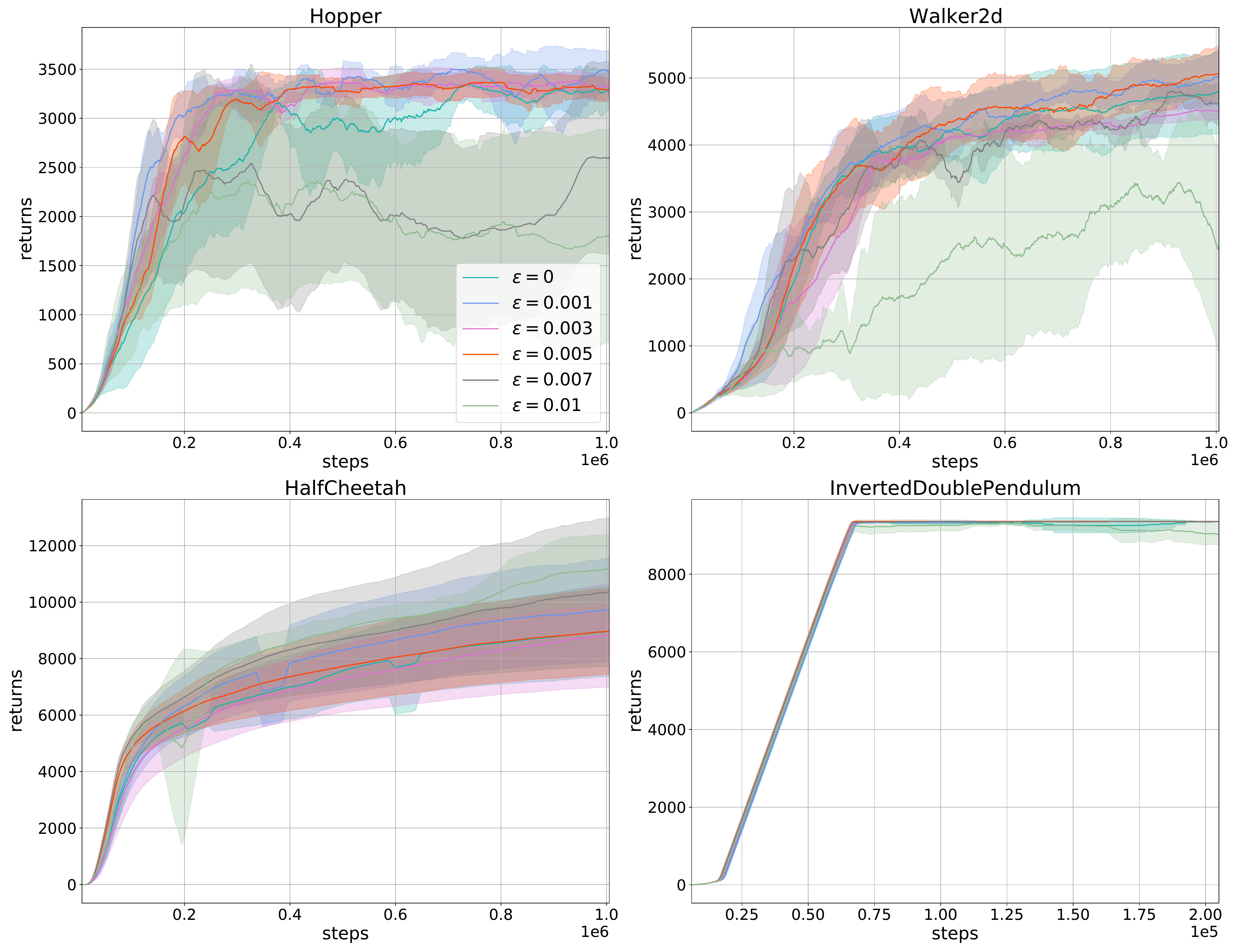}
\caption{Training curves for SC-SAC with different $\epsilon$.
}
\label{fig: performance_2}
\end{figure}

\noindent\textbf{Truncated  Gaussian  Distribution for Environmental Parameters} We compare SC-SAC with SAC in target environments with mass and friction sampled from truncated Gaussian distributions in Section 6. Specifically, we let $p_{t} = p_{s} (1 + \sigma_p \xi)$, where $\xi$ is sampled from standard Gaussian distribution and truncated by $\pm3$, $p_s$ and $p_t$ are the parameters (mass or friction) in source and target environments, and $\sigma_p$ is the standard deviation for this parameter. 
Note that small perturbations on environmental parameters (mass and friction) do not degrade the performance much, while large perturbations on environmental parameters are less likely to occur in practice. 
Thus, we choose $\sigma_{p}$ to make the performance significantly degrade at $\xi=\pm3$ based on the results in Figure 3 in Section 6. 
We list the value of $\sigma_p$ for each parameter in different tasks in Table \ref{sigma} below.

\begin{table}[h]
\centering 
\caption{The value of $\sigma_p$ for mass and friction in different tasks.}
\label{sigma}
\begin{tabular}{ccc}
\hline
& \textbf{$\sigma_p$-mass} & \textbf{$\sigma_p$-friction} \\ \hline
\textbf{Hopper}                                                            &  0.2    &   0.1    \\
\textbf{Walker2d}                                                          &  0.1  &  0.2  \\
\textbf{HalfCheetah}                                                       &  0.1  &   0.3  \\
\textbf{InvertedDoublePendulum} & 0.2 &   0.3  \\ 
\hline
\end{tabular}
\end{table}

\section{Additional Experimental Results} 
\setcounter{figure}{8}

\subsection{Comparative Evaluation}

\textbf{Comparison on Computational Efficiency} To compare the computational efficiency between SC-SAC and SAC, we run SC-SAC and SAC in four different tasks and report the mean time cost with standard deviation for training $1000$ steps in Table \ref{table: time_cost}. Results show that SC-SAC requires about $32.3\%$ additional time cost to calculate the gradient regularizer during training.

\begin{table}[h]
\centering 
\caption{Compare the computational efficiency between SC-SAC and SAC. We run each experiments with one GPU (Nvidia Geforce RTX 2080Ti) and report the wall-clock time cost (second) for $1000$-step training. }
\begin{tabular}{ccc}
\hline
                                                             & \textbf{SAC} & \textbf{SC-SAC} \\ \hline
\textbf{Hopper}                                                            &  $17.34\pm 0.44$    &   $22.75\pm 0.86$    \\
\textbf{Walker2d}                                                          &  $ 17.33\pm 0.37 $   &   $22.96 \pm 0.96 $  \\
\textbf{HalfCheetah}                                                       &  $17.74\pm 0.39 $  &   $ 23.58\pm 0.41 $  \\
\textbf{InvertedDoublePendulum} & $ 17.19\pm 0.40 $ &   $ 22.86\pm 0.45 $  \\ \hline
\end{tabular}
\label{table: time_cost}
\end{table}

\noindent\textbf{Return Distribution with  Truncated Gaussian Parameters} We report the return distribution of SAC and SC-SAC with truncated Gaussian distributed mass and friction. As shown in Figure \ref{fig: distribution2}, the return distribution of SC-SAC in the Hopper task is more concentrated in the high-value areas than SAC. 
Additionally, we find the return distributions of SC-SAC and SAC in the InvertedDoublePendulum task are similar, as performance degradation in this task only appears when the perturbations on environmental parameters are extremely large, which is less likely to occur in truncated Gaussian distributions.

\subsection{Ablation Study}

\textbf{Policy Evaluation v.s. Policy Improvement} We provide additional ablation study in the HalfCheetah and the InvertedDoublePendulum tasks to analyze the effects of the policy evaluation step and the policy improvement step in SC-SAC. Results in Figure \ref{fig: eval_impr_2} show that both the evaluation step and the improvement step play an important role in SC-SAC during training.

\noindent\textbf{Sensitivity Analysis on $\epsilon$} We provide additional sensitivity analysis on $\epsilon$ in the InvertedDoublePendulum and the HalfCheetah. Results in Figure \ref{fig: ablation2} show conclusions similar to that in Section 6.2. 
Then, we report the training curves with different $\epsilon$ in Figure \ref{fig: performance_2}. Results show that SC-SAC does not degrade the performance of SAC in source environments if $\epsilon$ is not too large.


\bibliography{aaai22}

\begin{thebibliography}{22}
\providecommand{\natexlab}[1]{#1}

\bibitem[{Abdullah et~al.(2019)Abdullah, Ren, Ammar, Milenkovic, Luo, Zhang,
  and Wang}]{abdullah2019wasserstein}
Abdullah, M.~A.; Ren, H.; Ammar, H.~B.; Milenkovic, V.; Luo, R.; Zhang, M.; and
  Wang, J. 2019.
\newblock Wasserstein robust reinforcement learning.
\newblock \emph{arXiv preprint arXiv:1907.13196}.

\bibitem[{Blanchet and Murthy(2019)}]{blanchet2019quantifying}
Blanchet, J.~H.; and Murthy, K. R.~A. 2019.
\newblock Quantifying Distributional Model Risk via Optimal Transport.
\newblock \emph{Math. Oper. Res.}, 44(2): 565--600.

\bibitem[{Boyd and Vandenberghe(2004)}]{boyd2004convex}
Boyd, S.; and Vandenberghe, L. 2004.
\newblock \emph{Convex optimization}.
\newblock Cambridge university press.

\bibitem[{Fujimoto, van Hoof, and Meger(2018)}]{td3}
Fujimoto, S.; van Hoof, H.; and Meger, D. 2018.
\newblock Addressing function approximation error in actor-critic methods.
\newblock In Dy, J.; and Krause, A., eds., \emph{Proceedings of the 35th
  International Conference on Machine Learning}, volume~80 of \emph{Proceedings
  of Machine Learning Research}, 1587--1596. PMLR.

\bibitem[{Haarnoja et~al.(2018)Haarnoja, Zhou, Hartikainen, Tucker, Ha, Tan,
  Kumar, Zhu, Gupta, Abbeel et~al.}]{sac}
Haarnoja, T.; Zhou, A.; Hartikainen, K.; Tucker, G.; Ha, S.; Tan, J.; Kumar,
  V.; Zhu, H.; Gupta, A.; Abbeel, P.; et~al. 2018.
\newblock Soft actor-critic algorithms and applications.
\newblock \emph{arXiv preprint arXiv:1812.05905}.

\bibitem[{Hou et~al.(2020)Hou, Pang, Hong, Lan, Ma, and Yin}]{hou2020robust}
Hou, L.; Pang, L.; Hong, X.; Lan, Y.; Ma, Z.; and Yin, D. 2020.
\newblock Robust reinforcement learning with Wasserstein constraint.
\newblock \emph{arXiv preprint arXiv:2006.00945}.

\bibitem[{Iyengar(2005)}]{robust-dp}
Iyengar, G.~N. 2005.
\newblock Robust Dynamic Programming.
\newblock \emph{Math. Oper. Res.}, 30(2): 257--280.

\bibitem[{Jiang et~al.(2021)Jiang, Li, Dai, Zou, and Xiong}]{mrpo}
Jiang, Y.; Li, C.; Dai, W.; Zou, J.; and Xiong, H. 2021.
\newblock Monotonic robust policy optimization with model discrepancy.
\newblock In Meila, M.; and Zhang, T., eds., \emph{Proceedings of the 38th
  International Conference on Machine Learning}, volume 139 of
  \emph{Proceedings of Machine Learning Research}, 4951--4960. PMLR.

\bibitem[{Lillicrap et~al.(2016)Lillicrap, Hunt, Pritzel, Heess, Erez, Tassa,
  Silver, and Wierstra}]{ddpg}
Lillicrap, T.~P.; Hunt, J.~J.; Pritzel, A.; Heess, N.; Erez, T.; Tassa, Y.;
  Silver, D.; and Wierstra, D. 2016.
\newblock Continuous control with deep reinforcement learning.
\newblock In \emph{International Conference on Learning Representations}.

\bibitem[{Mandlekar et~al.(2017)Mandlekar, Zhu, Garg, Fei-Fei, and
  Savarese}]{fgsm-rl}
Mandlekar, A.; Zhu, Y.; Garg, A.; Fei-Fei, L.; and Savarese, S. 2017.
\newblock Adversarially robust policy learning: Active construction of
  physically-plausible perturbations.
\newblock In \emph{2017 IEEE/RSJ International Conference on Intelligent Robots
  and Systems (IROS)}, 3932--3939.

\bibitem[{Mankowitz et~al.(2020)Mankowitz, Levine, Jeong, Abdolmaleki,
  Springenberg, Shi, Kay, Hester, Mann, and
  Riedmiller}]{model-misspecification}
Mankowitz, D.~J.; Levine, N.; Jeong, R.; Abdolmaleki, A.; Springenberg, J.~T.;
  Shi, Y.; Kay, J.; Hester, T.; Mann, T.; and Riedmiller, M. 2020.
\newblock Robust reinforcement learning for continuous control with model
  misspecification.
\newblock In \emph{International Conference on Learning Representations}.

\bibitem[{Mnih et~al.(2015)Mnih, Kavukcuoglu, Silver, Rusu, Veness, Bellemare,
  Graves, Riedmiller, Fidjeland, Ostrovski, Petersen, Beattie, Sadik,
  Antonoglou, King, Kumaran, Wierstra, Legg, and Hassabis}]{dqn}
Mnih, V.; Kavukcuoglu, K.; Silver, D.; Rusu, A.~A.; Veness, J.; Bellemare,
  M.~G.; Graves, A.; Riedmiller, M.~A.; Fidjeland, A.; Ostrovski, G.; Petersen,
  S.; Beattie, C.; Sadik, A.; Antonoglou, I.; King, H.; Kumaran, D.; Wierstra,
  D.; Legg, S.; and Hassabis, D. 2015.
\newblock Human-level control through deep reinforcement learning.
\newblock \emph{Nature}, 518(7540): 529--533.

\bibitem[{Nilim and Ghaoui(2004)}]{robust-mdp}
Nilim, A.; and Ghaoui, L. 2004.
\newblock Robustness in Markov Decision Problems with Uncertain Transition
  Matrices.
\newblock In Thrun, S.; Saul, L.; and Sch\"{o}lkopf, B., eds., \emph{Advances
  in Neural Information Processing Systems}, volume~16. MIT Press.

\bibitem[{OpenAI et~al.(2019)OpenAI, Akkaya, Andrychowicz, Chociej, Litwin,
  McGrew, Petron, Paino, Plappert, Powell, Ribas, Schneider, Tezak, Tworek,
  Welinder, Weng, Yuan, Zaremba, and Zhang}]{openai-cube}
OpenAI; Akkaya, I.; Andrychowicz, M.; Chociej, M.; Litwin, M.; McGrew, B.;
  Petron, A.; Paino, A.; Plappert, M.; Powell, G.; Ribas, R.; Schneider, J.;
  Tezak, N.; Tworek, J.; Welinder, P.; Weng, L.; Yuan, Q.; Zaremba, W.; and
  Zhang, L. 2019.
\newblock Solving Rubik's cube with a robot hand.
\newblock \emph{arXiv preprint}.

\bibitem[{Packer et~al.(2018)Packer, Gao, Kos, Kr{\"a}henb{\"u}hl, Koltun, and
  Song}]{benchmarks}
Packer, C.; Gao, K.; Kos, J.; Kr{\"a}henb{\"u}hl, P.; Koltun, V.; and Song, D.
  2018.
\newblock Assessing generalization in deep reinforcement learning.
\newblock \emph{arXiv preprint arXiv:1810.12282}.

\bibitem[{Pinto et~al.(2017)Pinto, Davidson, Sukthankar, and Gupta}]{rarl}
Pinto, L.; Davidson, J.; Sukthankar, R.; and Gupta, A. 2017.
\newblock Robust adversarial reinforcement learning.
\newblock In Precup, D.; and Teh, Y.~W., eds., \emph{Proceedings of the 34th
  International Conference on Machine Learning}, volume~70 of \emph{Proceedings
  of Machine Learning Research}, 2817--2826. PMLR.

\bibitem[{Rajeswaran et~al.(2017)Rajeswaran, Ghotra, Ravindran, and
  Levine}]{epopt}
Rajeswaran, A.; Ghotra, S.; Ravindran, B.; and Levine, S. 2017.
\newblock EPOpt: Learning Robust Neural Network Policies Using Model Ensembles.
\newblock In \emph{International Conference on Learning Representations}.
  OpenReview.net.

\bibitem[{Sutton and Barto(2018)}]{sutton2018reinforcement}
Sutton, R.~S.; and Barto, A.~G. 2018.
\newblock \emph{Reinforcement learning: An introduction}.
\newblock MIT press.

\bibitem[{Tessler, Efroni, and Mannor(2019)}]{tessler2019action}
Tessler, C.; Efroni, Y.; and Mannor, S. 2019.
\newblock Action robust reinforcement learning and applications in continuous
  control.
\newblock In Chaudhuri, K.; and Salakhutdinov, R., eds., \emph{Proceedings of
  the 36th International Conference on Machine Learning}, volume~97 of
  \emph{Proceedings of Machine Learning Research}, 6215--6224. PMLR.

\bibitem[{Tobin et~al.(2017)Tobin, Fong, Ray, Schneider, Zaremba, and
  Abbeel}]{dr}
Tobin, J.; Fong, R.; Ray, A.; Schneider, J.; Zaremba, W.; and Abbeel, P. 2017.
\newblock Domain randomization for transferring deep neural networks from
  simulation to the real world.
\newblock In \emph{2017 IEEE/RSJ International Conference on Intelligent Robots
  and Systems (IROS)}, 23--30.

\bibitem[{Todorov, Erez, and Tassa(2012)}]{mujoco}
Todorov, E.; Erez, T.; and Tassa, Y. 2012.
\newblock MuJoCo: A physics engine for model-based control.
\newblock In \emph{2012 IEEE/RSJ International Conference on Intelligent Robots
  and Systems}, 5026--5033.

\bibitem[{Zhang et~al.(2020)Zhang, Chen, Xiao, Li, Liu, Boning, and
  Hsieh}]{robust-observation}
Zhang, H.; Chen, H.; Xiao, C.; Li, B.; Liu, M.; Boning, D.; and Hsieh, C.-J.
  2020.
\newblock Robust Deep Reinforcement Learning against Adversarial Perturbations
  on State Observations.
\newblock In Larochelle, H.; Ranzato, M.; Hadsell, R.; Balcan, M.~F.; and Lin,
  H., eds., \emph{Advances in Neural Information Processing Systems},
  volume~33, 21024--21037. Curran Associates, Inc.

\end{thebibliography}
\bibliographystyle{aaai22}

\end{document}